\documentclass[final,5p,times,twopage,twocolumn,a4paper]{elsarticle}
\usepackage{hyperref}
\usepackage{graphicx}  
\usepackage{amsmath}
\usepackage{url}
\usepackage{adjustbox}
\usepackage{float}
\usepackage{multicol}
\usepackage{multirow}
\usepackage{rotating}
\usepackage{acronym} 
\usepackage{algorithm}
\usepackage{algpseudocode}
\usepackage{xcolor}
\usepackage{geometry}
\usepackage{setspace}
\usepackage{tabularx}
\usepackage{amsmath}
\usepackage{enumitem}
\usepackage{lipsum}
\usepackage{booktabs} % Load once
\usepackage{array}
\usepackage{fancyvrb}
\usepackage[skip=1em]{parskip}
\usepackage{threeparttable}
\usepackage{float}
\usepackage{xurl}
\usepackage{underscore}
\usepackage{textcomp}
\usepackage{framed}
\usepackage[T1]{fontenc}
\usepackage{lmodern}

\begin{document}
\begin{frontmatter}
\journal{arXiv}
 % ------ Line stretch -------------------------------------------------------
\title{MimirRAG: A Multi-Agent RAG Framework for Financial Data Retrieval with Metadata Integration}

\author[label1]{Magnus Samuelsen}
\ead{masa23ap@student.cbs.dk}

\author[label1]{Wilmer Nyström}
\ead{winy23ab@student.cbs.dk}

\author[label1]{Somnath Mazumdar}
\ead{sma.digi@cbs.dk}

\author[label2]{Mansoor Hussain}
\ead{mahu@jyskebank.dk}

\author[label2]{Mikkel Strange}
\ead{mikkel.strange@jyskebank.dk}

\affiliation[label1]{organization={Copenhagen Business School},
            addressline={Solbjerg Plads 3},
            city={Frederiksberg},
            postcode={2000},
            country={Denmark}}

\affiliation[label2]{organization={Jyske Bank},
            city={Copenhagen},
            country={Denmark}}

\begin{abstract}
Retrieval-augmented generation (RAG) systems offer a promising approach to reduce hallucinations and improve answer accuracy in large language models (LLMs), a requirement for reliable, financial analysis where answers must be grounded in verifiable evidence from filings rather than generated from model priors. However, designing RAG systems that extract meaningful insights from mixed financial documents and integrate into analyst workflows remains challenging. This paper introduces MimirRAG (Metadata-Integrated Multi-Agent Information Retrieval), a multi-agent RAG system developed iteratively to address these challenges. MimirRAG features a modular pipeline encompassing structure-preserving parsing of PDF filings, table-aware chunking, metadata extraction, agent-based retrieval with query planning and hybrid search, validation, and context-aware generation with numerical reasoning support. Our ablation study identifies three key technical enablers for effective financial RAG: metadata integration, table-aware chunking, and an agentic workflow. MimirRAG was evaluated quantitatively using FinanceBench and qualitatively through expert validation with four financial analysts. The system achieved 89.3\% accuracy on FinanceBench, outperforming the original benchmark baselines. Expert feedback highlighted that successful deployment also requires calibrated trust, comprehensive data integration, and user personalization. We conclude that combining multi-agent RAG architecture with human-centric design principles can improve the extraction of meaningful insights in financial analysis.
\end{abstract}

\begin{keyword}
RAG \sep  GPT \sep  Financial Analysis \sep  Agent \sep  Metadata \sep  Retrieval
\end{keyword}

\end{frontmatter}

%%%%%%%%%%%%%%%%%%%%%%%%%%%%%%%%%
\section{Introduction}
\label{chap:introduction}
%%%%%%%%%%%%%%%%%%%%%%%%%%%%%%%%%
Large language models (LLMs)\footnote{Generative Pre-trained Transformer (GPT) is a specific type of LLM developed by OpenAI.} have begun to transform workflows in finance. The financial sector, with its large volume of reports, filings, and other analyst-facing documents, stands to benefit from these advances~\cite{nie2024survey}. However, standard LLMs in financial analysis carry the risk of \textit{hallucinations}, unsupported or factually incorrect statements~\cite{ray_srinivasan_2024amazon,sarmah2023towards}. In finance, even minor factual errors can have reputational, monetary, or legal consequences, making hallucination mitigation a key deployment challenge~\cite{nie2024survey}. Furthermore, the volume of financial reports often leads to information overload, hindering analyst efficiency and prediction accuracy, especially for less experienced analysts~\cite{impink2022regulation}. Retrieval-augmented generation (RAG) offers a solution by connecting LLMs to external knowledge sources: RAG retrieves relevant evidence for a query and supplies it as context to the generator, constraining answers to retrieved, citable evidence rather than relying solely on parametric memory~\cite{lewis2020retrieval,gao2023retrieval}. This focus on accuracy and traceability makes RAG attractive for knowledge-intensive domains like finance~\cite{sarmah2023towards}, with leading financial institutions developing RAG systems such as AskResearchGPT\footnote{https://www.morganstanley.com/press-releases/morgan-stanley-research-announces-askresearchgpt} to assist analysts in querying research report libraries.

\textbf{Problem Context:}~Academia has increasingly turned its attention to customizing RAG systems for financial applications. Most existing work emphasizes technical performance improvements such as optimizing retrieval through enhanced embedding models~\cite{iaroshev2024evaluating}, refining chunking strategies~\cite{yepes2024financial}, or extending RAG architectures to better align with analyst workflows~\cite{zhou2024finrobot}. However, a critical dimension remains underexplored: the financial analysts' perspective. There is limited understanding of how these intended end-users perceive, interact with, and integrate such systems into their daily workflows. This paper addresses this gap by investigating both the technical design of RAG systems for finance and their human-centered evaluation.

\textbf{Research Context:}~We focus on the strategic role of metadata. While its potential to enhance RAG performance is acknowledged, systematic evaluation of metadata as a foundational component in financial RAG systems remains rare. Our primary research question is:~\textbf{How can RAG-based solutions extract meaningful financial insights?} This is explored through three sub-questions: \textit{1)} How can a RAG-based solution be designed for financial information search tasks? \textit{2)} What design principles improve RAG systems' precision and usability for financial analysts? \textit{3)} What key factors are crucial for integrating RAG-based solutions into financial analysts' workflows?

To answer these we designed and prototyped~\textit{MimirRAG}, with a user interface (UI) for human evaluation. The target task is open-book financial question answering over corporate filings, including fact lookup, cross-period comparison, ratio calculation, and reasoning over tabular disclosures. Our corpus consists of PDF filings, mixed semi-structured documents containing free text, section hierarchies, and visually rendered tables. The system converts these into a structure-preserving intermediate representation before chunking and retrieval. Performance is assessed using FinanceBench~\cite{islam2023financebench}, a benchmark tailored for open-book financial Q\&A used in both academic~\cite{lee2024multireranker,setty2024improving} and industry settings~\cite{Leng2024LongCRA,rafiq2024ragie,rafiq2024ragie2,vectify2025mafingithub}. This quantitative evaluation is supplemented by qualitative insights from four expert analysts. Our key contributions are:
\begin{enumerate}
    \item We present a five-agent modular RAG architecture called \textit{MimirRAG}, structured around pre-retrieval, retrieval, generation, and orchestration stages. It achieves 89.3\% accuracy on FinanceBench through metadata integration and multi-agent design. Our ablation study demonstrates how structured document metadata (company name, report type, date) enhances retrieval precision.
    
   \item Our investigation, combined with qualitative feedback from four financial analysts, reveals five design principles for improving precision and usability: \textit{i)}~domain-appropriate chunking strategies\footnote{We use `document' to refer to the source filing, and `chunk' to refer to chunks retrieved by the retrieval system.}, \textit{ii)}~metadata-driven targeted retrieval, \textit{iii)}~layered retrieval with granular filtering and validation, \textit{iv)}~calibrated trust through transparency and consistency, and \textit{v)}~integration with existing analyst data ecosystems and workflows.
   
   \item We find that integrating RAG tools into financial analysts' workflows depends on three factors beyond technical performance: \textit{i)}~calibrated trust, \textit{ii)}~comprehensive data integration, and \textit{iii)}~personalized augmentation of expertise.  
\end{enumerate}
%
%%%%%%%%%%%%%%%%%%%%%%%%%%%
    \section{Related Work}
    \label{chap:literature}
    %%%%%%%%%%%%%%%%%%%%%%%%%%%%
    We survey the progression from early fine-tuned models such as FinBERT~\cite{araci2019finbert} to modular RAG pipelines that combine chunking, metadata filtering, and agentic frameworks. We identify three gaps in current financial RAG research:
    
    \textit{1)~Metadata:} Financial filings possess rich structured information (e.g., company, date, fiscal year, report type), yet systematic leverage of this metadata throughout the RAG pipeline remains partial or underexplored. \textit{2)~Practical Insights:} Benchmark performance dominates the literature, qualitative feedback from financial analysts remains scarce.~\textit{3)~End-to-end Evaluation:} Comprehensive testing on realistic financial benchmarks assessing the complete retrieval-to-generation pipeline is less common, many studies focus on component-level optimizations or retrieval metrics alone.

\subsection{LLMs in Finance}
Early financial natural language processing (NLP) saw fine-tuned models like FinBERT enhance BERT by training on domain-specific texts. Subsequent developments include proprietary models like BloombergGPT~\cite{wu2023bloombergGPT} and open-source initiatives such as FinGPT~\cite{liu2023fingpt} and the multi-agent FinRobot~\cite{yang2024finrobot}, all underscoring the value of domain adaptation. RAG systems extend this principle by grounding LLM responses in specific, externally retrieved contextual documents.

\subsection{RAG in Financial Analysis}
Research in financial RAG has rapidly evolved, exploring various pipeline enhancements (see Table~\ref{tab:rag_finance_review} for a summary of key systems and their limitations). Initial efforts were focused to optimize core RAG components. For instance, studies have investigated the impact of different embedding models~\cite{iaroshev2024evaluating} or developed domain-specific chunking strategies, (such as element-based chunking for financial documents~\cite{yepes2024financial}). Others have explored advanced retrieval techniques including query expansion and re-ranking~\cite{setty2024improving,lee2024multireranker}, multi-document retrieval using semantic tagging and knowledge graphs (KGs)~\cite{shah2024multi}, or hybrid approaches combining vector search with KGs~\cite{sarmah2024hybridrag}. Some systems integrate RAG with chain-of-thought prompting and fine-tuning for tasks like stock prediction and question answering~\cite{li2024alphafin}.

Despite these advancements, several challenges persist. Many approaches struggle with extracting and reasoning over structured data such as tables and performing numerical calculations~\cite{iaroshev2024evaluating,setty2024improving,shah2024multi}, or they incur high computational costs~\cite{sarmah2024hybridrag,kim2025optimizingRag,li2024alphafin,lee2024multireranker}. While the importance of metadata has been noted (e.g., for document selection~\cite{lai2024sec} or filtering in specific contexts~\cite{sarmah2023towards}), a systematic investigation into its deep integration and impact across the entire RAG pipeline is often missing, with some studies not testing its effect or explicitly excluding it~\cite{kim2025optimizingRag,setty2024improving,sarmah2024hybridrag}. Furthermore, as highlighted by Kim et al.~\cite{kim2025optimizingRag} and evident in Table~\ref{tab:rag_finance_review}, comprehensive end-to-end evaluations incorporating human validation remain relatively uncommon.

\begin{table*}
\centering
\caption{Summary of RAG-based systems in financial applications}
\label{tab:rag_finance_review}
\begin{adjustbox}{width=0.85\textwidth}
\begin{tabularx}{\textwidth}{p{2.0cm} X X X X} 
\toprule
\textbf{Paper} & \textbf{Focus Area} & \textbf{Methodology} & \textbf{Evaluation Method} & \textbf{Key Limitation} \\
\midrule
Iaroshev et al.~\cite{iaroshev2024evaluating} & Financial QA & 
    \textbullet Testing various embeddings 
    & 
    \textbullet Context relevance \newline
    \textbullet Faithfulness \newline
    \textbullet Answer quality 
    & 
    \textbullet Poor on tables \newline
    \textbullet Numerical reasoning weak \newline
    \textbullet Format-specific issues \\
\midrule % Replaced \addlinespace
Kim et al.~\cite{kim2025optimizingRag} & Retrieval \& End‑to‑end RAG & 
    \textbullet Dense–sparse hybrid \newline
    \textbullet Rerank 
    & 
    7 QA sets (incl. FinanceBench, FinDER); NDCG@10, RAGAS 
    & 
    \textbullet High compute cost \newline
    \textbullet No metadata filtering \\
\midrule
Lai et al.~\cite{lai2024sec} & Financial QA Corpus; Multi-doc quantitative reasoning; Dynamic refresh. & 
    \textbullet SEC-QA framework (dynamic QA gen.) \newline
    \textbullet Program-of-thought RAG on financial filings. &
    \textbullet QA Accuracy \& Retrieval metrics for multi-doc quantitative QA. &
    \textbullet Assumes data-doc mapping. \newline
    \textbullet CodeGen higher compute cost. \\
\midrule % Replaced \addlinespace
Lee et al.~\cite{lee2024multireranker} & Pre-retrieval \& Reranking & 
    \textbullet Query expansion \newline
    \textbullet Long-context fusion 
    & 
    FinanceRAG metrics (accuracy, numerical reasoning) 
    & 
    \textbullet High computational cost \newline
    \textbullet Manual prompt tuning \\
\midrule % Replaced \addlinespace
Li et al.~\cite{li2024alphafin} & QA + Stock Prediction & 
    \textbullet CoT prompting \newline
    \textbullet Real-time financial data 
    & 
    ARR, Accuracy (vs FinGPT) 
    & 
    \textbullet Computational expensive at large scales \\ 
\midrule 
Sarmah et al.~\cite{sarmah2024hybridrag} & Hybrid RAG & 
    \textbullet VectorRAG \newline
    \textbullet GraphRAG \newline
    \textbullet Metadata filtering 
    & 
    \textbullet Context precision \newline
    \textbullet Faithfulness \newline
    \textbullet Answer relevance 
    & 
    \textbullet High compute cost \newline
    \textbullet Manual KG creation \newline
    \textbullet Metadata effect not tested \\
\midrule % Replaced \addlinespace

Sarmah et al.~\cite{sarmah2023towards} & Reducing hallucination in financial report QA extraction. & 
    \textbullet RAG (PaLM2, Llama2, etc.) \newline
    \textbullet Metadata filtering for targeted retrieval \newline
    \textbullet Chunking; MMR.
    & 
    \textbullet BERTScore, BARTScore, Jaro.
    & 
    \textbullet Metadata benefit is less clear for some metrics. \newline
    \textbullet Outdated models. \\
\midrule % Replaced \addlinespace

Setty et al.~\cite{setty2024improving} & Retrieval Optimization & 
    \textbullet Query expansion \newline
    \textbullet Re-ranking \newline
    \textbullet Chunking 
    & 
    FinanceBench (retrieval + QA metrics) 
    & 
    \textbullet Multi-doc still weak \newline
    \textbullet Poor on structured data \newline
    \textbullet No results on Metadata \\
\midrule % Replaced \addlinespace

Shah et al.~\cite{shah2024multi} & Multi-document RAG & 
    \textbullet KG-RAG \newline
    \textbullet Semantic tagging for multi-source QA 
    & 
    9 metrics incl.: \newline
    \textbullet Faithfulness \newline
    \textbullet Correctness 
    & 
    \textbullet Graph errors \newline
    \textbullet Table/chart parsing weak \newline
    \textbullet Tagging dependency \\
\midrule % Replaced \addlinespace
Taghvaei et al.~\cite{Taghvaei2024CombiningFDA} & Stock Price Prediction; Multi-modal RAG (10-K \& news). & 
    \textbullet LLM classifier with RAG for news (metadata filtered). \newline
    \textbullet Zero/few-shot. & 
    \textbullet 3/6-month stock movement prediction (Weighted F1, MCC). & 
    \textbullet Few-shot limited by prompt length/token limits. \\
\midrule % Replaced \addlinespace

Yepes et al.~\cite{yepes2024financial} & Chunking Strategies & 
    Element-based chunking (headers/tables) 
    & 
    FinanceBench Q/A performance 
    & 
    Not a full system test \\ 
\bottomrule
\end{tabularx}
\end{adjustbox}
\end{table*}

\subsection{Modular RAG}
The trend towards modular RAG architectures decomposes the standard pipeline into specialized, interconnected components~\cite{gao2023retrieval,singh2025agentic}. In financial applications, these systems often adopt an agentic approach, aiming to replicate the analytical workflow of human analysts through collaborative AI agents~\cite{chen2024moa,yang2024finrobot,zhou2024finrobot} rather than serving solely as passive tools. For example, Chen et al.~\cite{chen2024moa} utilized a mixture-of-agents framework with small, specialized LLMs for discrete tasks. Systems like FinRobot~\cite{yang2024finrobot,zhou2024finrobot} employ multiple agents, often with chain-of-thought prompting and RAG-based grounding, for complex tasks such as equity research. While promising for enhanced reasoning and flexibility, these agentic systems can be computationally intensive and often lack proper benchmarking or evaluation within real-world analyst workflows.

\subsection{Contextualization} 
Our work addresses three limitations in financial RAG research. First, while prior studies incorporate metadata for filtering~\cite{sarmah2023towards} or document selection~\cite{lai2024sec}, its role as a systematically integrated component across the full RAG pipeline remains underexplored~\cite{poliakov2024metadata}. We examine how metadata integration influences both retrieval performance and overall system design.

Second, existing research largely focuses on optimizing isolated components with limited emphasis on end-to-end evaluation under realistic conditions. We conduct an evaluation of MimirRAG on FinanceBench, complemented by expert validation, to derive design principles for improving precision and usability.

Third, factors influencing practical adoption, trust calibration, perceived usefulness, and workflow integration, remain insufficiently understood. By involving financial analysts in evaluation, we identify key determinants for integrating RAG systems into real-world workflows.

%%%%%%%%%%%%%%%%%%%%%%%%%%%%%%%%%%
\section{Methodology}
\label{chap:methodology_framework}
%%%%%%%%%%%%%%%%%%%%%%%%%%%%%%%%%%
This research employed a structured and iterative methodological approach, integrating the conceptual guidance of Saunders et al.'s Research Onion~\cite{saunders2023researchframework} with the procedural framework of Action Design Research (ADR)~\cite{sein2011actiondesign}. The Research Onion guided our high-level methodological choices, leading us to adopt a critical realist research philosophy, an abductive research approach, and a mixed-methods strategy. This involved a case study at a major Danish bank to understand analyst information workflows and an experimental strategy for the iterative development and refinement of the MimirRAG prototype.

\begin{figure*}[!hbt]
\centering
\includegraphics[width=0.8\textwidth]{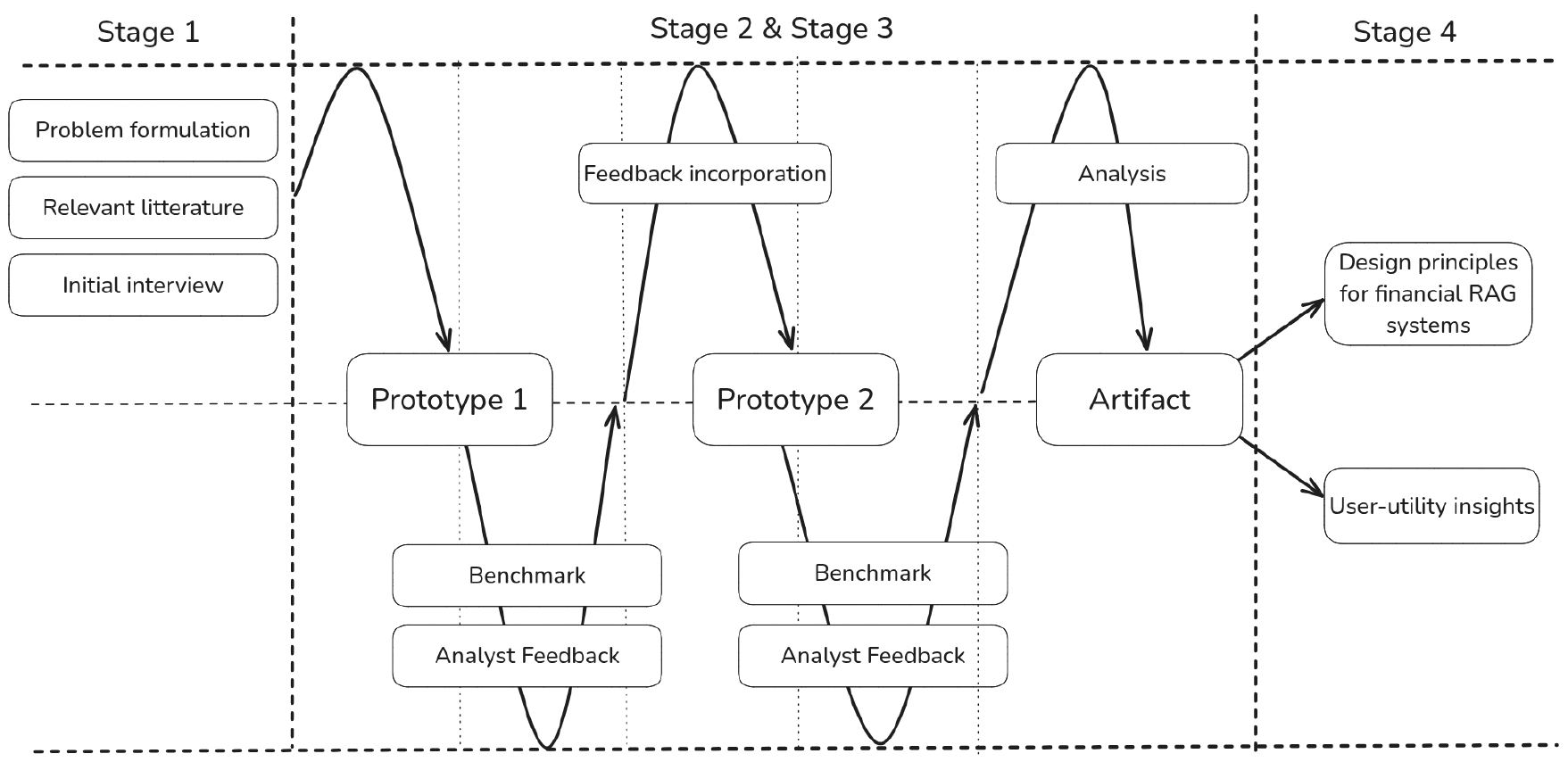}
\caption{Iterative BIE process in MimirRAG development.}
\label{fig:BIE_Overview_condensed}
\end{figure*}

The core of our research design was ADR, which conceptualizes the IT artifact as an \textit{ensemble artifact} shaped by its organizational context and the learning that occurs through its use~\cite{sein2011actiondesign}. ADR facilitates generalizable design knowledge through iterative cycles of Building, Intervention, and Evaluation (BIE) (refer to Figure~\ref{fig:BIE_Overview_condensed}), which are particularly beneficial for developing RAG systems where user interaction and contextual fit are important. Our research followed ADR's main stages.
\begin{itemize}
    \item \textit{Problem Formulation (Stage 1):} We initiated the research by identifying real-world analyst challenges (practice-inspired research) and designing an initial MimirRAG system based on established theories in RAG, information retrieval, and NLP (theory-ingrained artifact).
    \item \textit{Building, Intervention, and Evaluation (BIE Cycles - Stage 2):} MimirRAG was developed and iteratively refined through two BIE cycles. The artifact was deployed within the financial analysis environment of a major Danish bank. Four financial analysts interacted with the system (intervention), and their feedback directly informed its redesign (building), embodying the principle of reciprocal shaping. This involved mutually influential roles, with our team providing RAG expertise and analysts offering domain knowledge. Authentic and concurrent evaluation, integrating both benchmark testing and practitioner feedback, was continuous within each BIE cycle.
    \item \textit{Reflection and Learning (Stage 3):} Insights from benchmark performance (using FinanceBench) and expert validation were used to identify technical weaknesses (e.g., numerical reasoning, table parsing) and usability challenges, guiding priorities for subsequent development.
    \item \textit{Formalization of Learning (Stage 4):} The iterative process culminated in the derivation of practical design principles for financial RAG systems and theoretical insights into factors influencing their adoption (generalized outcomes).
\end{itemize}
This ADR process (Figure~\ref{fig:BIE_Overview_condensed}) facilitated a comprehensive evaluation of MimirRAG's technical performance and real-world applicability, addressing our main research question. The design of the RAG solution for financial information search (first subresearch question) was primarily addressed during ADR Stages 1 \& 2 through initial design, iterative development, and technical benchmarking. Design principles improving RAG precision and usability for financial analysts (second subresearch question) were formalized in ADR Stage 4, drawing from consolidated benchmark and analyst insights. Key factors for integrating RAG solutions into analysts' workflows (third subresearch question) were identified through analyzing analyst feedback and usability from ADR Stages 2 \& 3.

\subsection{Analyst Feedback and Evaluation}
A key ADR component, and addressing a noted gap in financial RAG research, was the deep involvement of end-users. We conducted two iterative expert validation cycles with four financial analysts from a major Danish bank, as summarized in Table~\ref{tab:expert_validation_overview}.

\begin{table*}[!hbt]
\centering
\caption{Overview of Expert Validation Activities and Participant Involvement Across Iterations}
\label{tab:expert_validation_overview}
\begin{adjustbox}{width=.8\textwidth}
\begin{tabular}{|c|c|c|c|}
\hline
\textbf{Evaluation Activity}       & \textbf{Brief Description}                                                                                                 & \textbf{Participants}                                     & \textbf{Iteration}                                             \\ \hline
Scoping Interview                  & \begin{tabular}[c]{@{}c@{}}Understand P1 workflow \\ \& select Iter. 1 docs.\end{tabular}                                  & P1                                                        & \begin{tabular}[c]{@{}c@{}}Setup \\ (Pre-Iter. 1)\end{tabular} \\ \hline
In-depth Semi-Structured Interview & \begin{tabular}[c]{@{}c@{}}Assess baseline system: \\ usability, trust, value.\end{tabular}                                & P1                                                        & 1                                                              \\ \hline
A/B Survey                         & \begin{tabular}[c]{@{}c@{}}Compare MimirRAG vs. \\ Copilot outputs.\end{tabular}                                           & P1                                                        & 1                                                              \\ \hline
Email Feedback on Notes Feature    & \begin{tabular}[c]{@{}c@{}}Qualitative feedback on \\ `search-my-notes' feature.\end{tabular}                              & P1                                                        & 2                                                              \\ \hline
Semi-Structured Interviews         & \begin{tabular}[c]{@{}c@{}}Broader qualitative feedback: \\ usability, trust, improvements.\end{tabular}                   & \begin{tabular}[c]{@{}c@{}}P2, \\ P3, P4\end{tabular}     & 2                                                              \\ \hline
Analyst Feedback Survey            & \begin{tabular}[c]{@{}c@{}}Quantify perceptions (helpfulness, trust, \\ time-saving) \& gather open feedback.\end{tabular} & \begin{tabular}[c]{@{}c@{}}P1, P2, \\ P3, P4\end{tabular} & 2                                                              \\ \hline
\end{tabular}
\end{adjustbox}
\end{table*}

Participant 1 (P1), a senior sell-side analyst (Healthcare, more than 10 years experience), was central to Iteration 1. Initial scoping with P1 identified relevant documents (e.g., Novo Nordisk\footnote{Novo Nordisk A/S is a Danish multinational pharmaceutical company \url{https://www.novonordisk.com/}} annual reports, SEC Form 20-F filings) and workflow pain points, informing the Iteration 1 knowledge base and interface design. After a two-week trial, P1 participated in a semi-structured interview focusing on usability, trust, and improvements. Iteration 2 incorporated P1's feedback (notably private notes integration) and expanded evaluation to three additional analysts (P2: Director-level, P3: Assistant, P4: Senior) for diverse perspectives.

%%%%%%%%%%%%%%%%%%%%%%%%%%%%%%
\section{System Overview}
\label{chap:architecture}
\begin{figure*}[!hbt]
    \centering
    \includegraphics[width=0.8\textwidth]{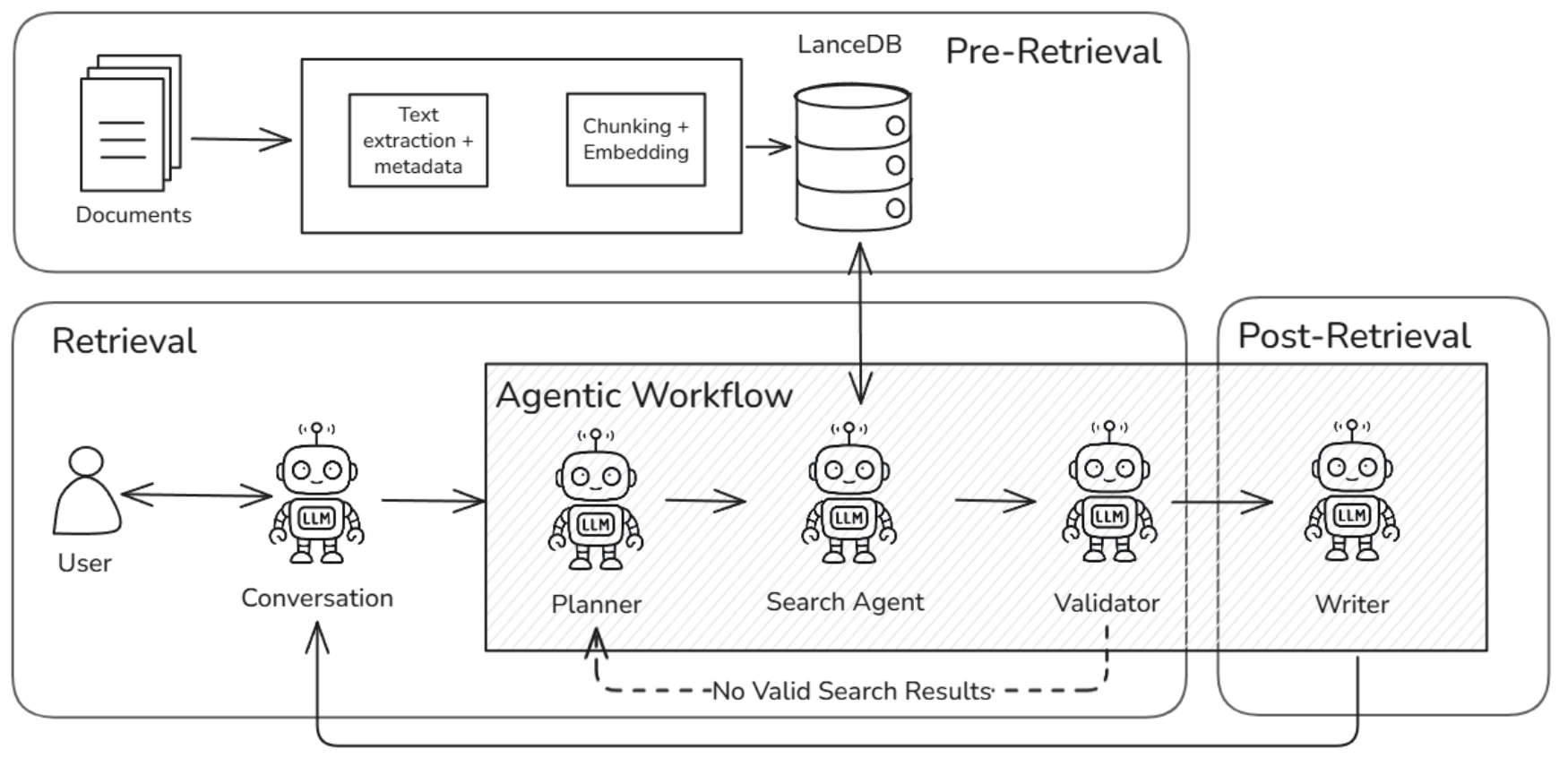}
    \caption{MimirRAG pipeline: pre-retrieval, retrieval, and post-retrieval stages.}
    \label{fig:system_arch}
\end{figure*}

%%%%%%%%%%%%%%%%%%%%%%%%%%%%%%
MimirRAG (\textbf{M}etadata-\textbf{I}ntegrated \textbf{M}ulti-Agent \textbf{I}nformation \textbf{R}etrieval) is a modular, agentic RAG workflow comprising five loosely coupled agents: Extractor, Planner, Search, Validator, and Writer. Each agent executes a distinct sub-task in the RAG pipeline and communicate via structured text-based outputs, enabling modularity and supporting independent adaptation or replacement. MimirRAG separates the pipeline into three logical stages: pre-retrieval, retrieval, and post-retrieval (see Figure~\ref{fig:system_arch}) following best practices in RAG systems~\cite{gao2023retrieval,singh2025agentic}.

Agentic decomposition provides several advantages. For example, failed retrievals can be dynamically retried with adaptive query reformulation; individual agent logic, prompts, or models can be tuned or swapped in isolation; and heterogeneous models (e.g., lightweight search, heavyweight generation) can be efficiently composed. This modular architecture further facilitates rapid integration of new workflows, tools, or agent capabilities with minimal system-wide impact.
%

%========================================%
\subsection{Pre-Retrieval Pipeline}
\label{sec:Pre-retrival-Pipeline}
\begin{figure*}[!hbt]
    \centering
    \includegraphics[width=0.8\textwidth]{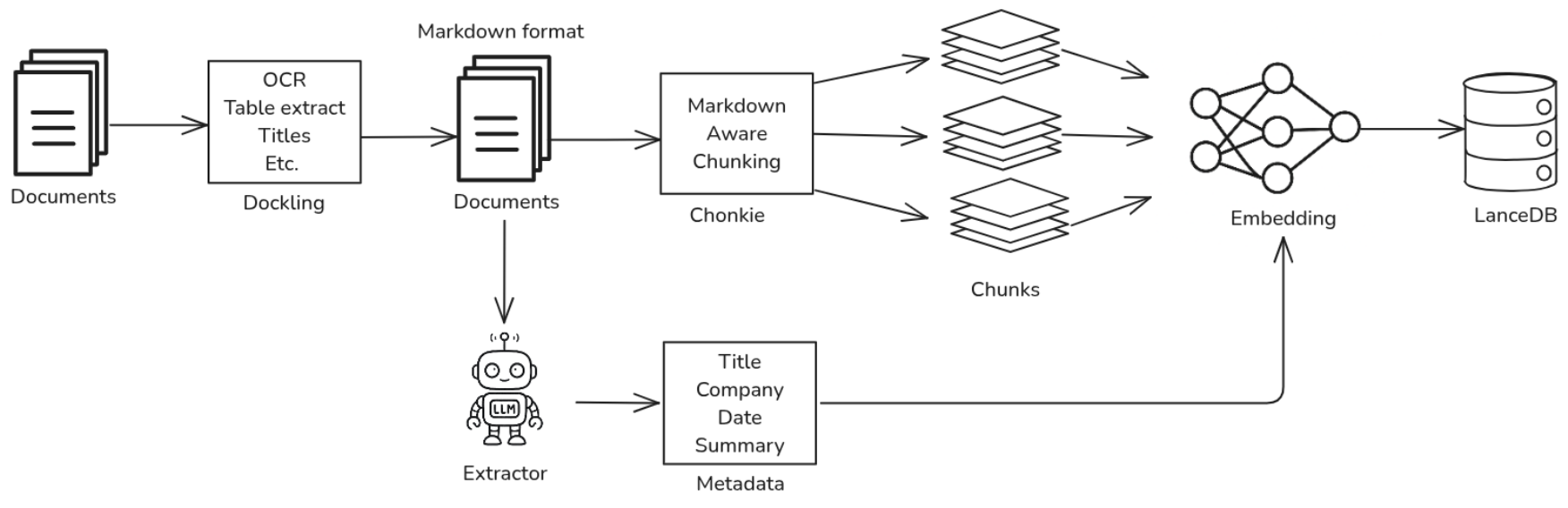}
    \caption{Pre-retrieval data pipeline representation}
    \label{fig:data_pipeline}
\end{figure*}

%========================================%
The pre-retrieval stage transforms raw financial documents into a structured corpus with metadata and embeddings to enable hybrid-based (Best Match 25 (BM25)\footnote{It is a ranking function that extends TF-IDF (term frequency-inverse document frequency) by considering term frequency saturation and document length.} + vector) search. As illustrated in Figure~\ref{fig:data_pipeline}, the pipeline includes: \textit{1)} robust document parsing and chunking, \textit{2)} structured metadata extraction, and \textit{3)} embedding and indexing in LanceDB.

\subsubsection{Document Parsing and Chunking}  
Raw PDFs are processed using \textsc{Docling}~\cite{auer2024docling}\footnote{\url{https://github.com/docling-project/docling}}, integrating DocLayNet~\cite{pfitzmann2022doclaynet} for layout analysis, TableFormer~\cite{nassar2022tableformer} for table extraction, and optical character recognition (OCR) as needed. Here, the PDF is the source delivery format of the filing, while the filing content itself contains both prose and visually rendered tables. We treat the corpus as mixed semi-structured documents. Docling converts each PDF into a structured intermediate representation and then into Markdown while preserving headings, section boundaries, and table content where possible. This structural preservation is particularly valuable in finance, given documented challenges with tables and numerical data~\cite{iaroshev2024evaluating,setty2024improving,shah2024multi}. OCR and table extraction are not error-free, especially for scanned or image-heavy tables, and can introduce segmentation or cell-boundary errors; our downstream chunking and validation steps are designed to mitigate, but not eliminate, these failure modes.

Documents are normalized to Markdown, then recursively chunked using \textsc{Chonkie}\footnote{\url{https://github.com/chonkie-inc/chonkie}}, splitting on headings, paragraphs, and whitespace, with fixed chunk and overlap sizes ($L_\text{max}=1800$, $\Delta=300$), following industry best practices~\cite{merrick2024arctic}. In this representation, tables are serialized as Markdown table blocks rather than flattened into plain text. To reduce errors caused by split tables, consecutive chunks are merged when one chunk ends with a table row and the next begins with a table row, up to a maximum merged length of 3600 characters. This table-aware chunking reduces cases where a retrieved chunk contains only a partial set of table rows needed for a financial calculation.

\subsubsection{Extractor Agent}
Metadata extraction by the \textit{Extractor agent} leverages the first 1024 tokens (~2–3 pages) of each parsed document to obtain essential metadata (see Table~\ref{tab:metadata}), namely title, company name, keywords, summary, date, and report type. These metadata are created by an LLM-based extraction step over the document prefix because these fields are typically available near the front matter of annual reports and filings. Extracted company names are standardized via local FTS (full text search) matching in LanceDB, with fallback lookups through SEC EDGAR\footnote{The Electronic Data Gathering, Analysis, and Retrieval (EDGAR) system is used by the U.S. Securities and Exchange Commission (SEC) to collect, store, and provide public access to financial and operational filings submitted by companies and individuals.} and Danish central business register (CVR) APIs for robustness. Retrieved metadata is stored and indexed for ater use in hierarchical retrieval.
\begin{table*}[!hbt]
\centering
\caption{Structured metadata extracted by the Extractor agent}
\label{tab:metadata}
\begin{adjustbox}{width=.8\textwidth}
\begin{tabular}{|c|l|}
\hline
\textbf{Metadata} & \multicolumn{1}{c|}{\textbf{Description}}                                                                  \\ \hline
Title             & [Report Type] – [Year] ([Company Name])                                                                    \\ \hline
Company Name      & Standardized primary company name                                                                          \\ \hline
Keywords          & \multicolumn{1}{c|}{Up to five keywords capturing central financial topics, sectors, or strategic actions} \\ \hline
Summary           & Short summary of the input text                                                                            \\ \hline
Date              & Fiscal period-end date or publication date found in the document                                           \\ \hline
Report Type       & Specific document category (e.g., 10-K, Annual Report, Press Release)                                      \\ \hline
\end{tabular}
\end{adjustbox}
\end{table*}

Using the metadata taxonomy in Table~\ref{tab:metadata}, the Extractor agent improves document traceability and supports hierarchical retrieval by restricting the candidate document set before chunk-level search. In other words, metadata improves retrieval primarily by reducing the search space and thereby lowering semantic ambiguity during ranking. The same metadata are also passed forward as source context during final answer generation, as described in Section~\ref{sec:search_agent} and Section~\ref{sec:structured_input}.

\subsection{Embedding and Vector Storage}
Document-level embeddings are generated from the summaries produced by the Extractor agent, while chunk-level embeddings are generated directly from the chunk text. \texttt{snowflake-arctic-embed-m v2.0}~\cite{yu2024arctic}\footnote{\url{https://huggingface.co/Snowflake/snowflake-arctic-embed-m-v2.0}} embedding model (768 dimensions),  is selected for its  balance of embedding quality, inference speed, and strong performance on MTEB~\cite{enevoldsen2025mmtebmassivemultilingualtext} and FinMTEB~\cite{tang2025finmteb} benchmarks. These embeddings support semantic retrieval, while BM25 indexes over metadata and chunk text support sparse lexical retrieval; the combination enables the hybrid search setup used by the Search agent.

Embeddings and metadata are stored in LanceDB\footnote{\url{https://github.com/lancedb/lancedb}}, a serverless, embedded vector database. LanceDB utilizes the Lance open-source columnar format for efficient analytical queries and retrieval. In addition to vector search, LanceDB supports FTS indexes (BM25) on metadata fields (e.g., \texttt{title}, \texttt{summary}) and chunk content, which are crucial for the hybrid search capabilities detailed in Section~\ref{sec:search_agent}.

\subsection{Relational Document Linkage}
\begin{figure*}[!hbt]
    \centering
    \includegraphics[width=0.8\textwidth]{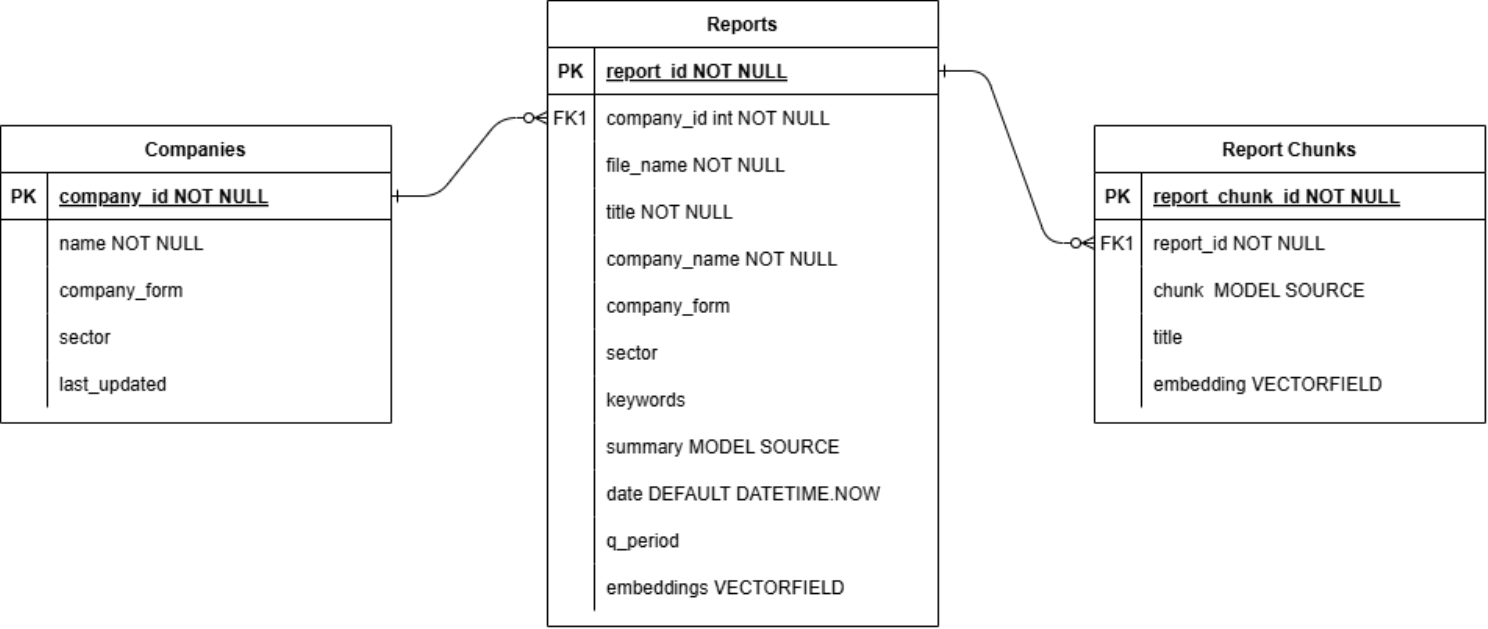}
    \caption{Entity-relationship (ER) diagram of vector database}
    \label{fig:ER}
\end{figure*}

Each chunk is linked to its source document using foreign key relationships (Figure~\ref{fig:ER}), ensuring retrieved chunks can be traced to their parent document. Document-level metadata and summaries extracted by the Extractor agent are stored with the parent document. This design supports explainability and our hierarchical retrieval approach.
%

%==================================
\subsection{Retrieval Pipeline}
%==================================
The retrieval stage begins once the user submits a query. Its goal is to return a concise, high-quality set of semantically aligned chunks with the query. Our system uses an agentic retrieval workflow including query decomposition, metadata filtering, hybrid search, and chunk validation.

%=====================================
\subsubsection{Planner Agent (Query Decomposition)}
\label{sec:planner_agent}
%=====================================

The Planner agent initiates the retrieval process. Upon receiving the user's  natural-language query, its primary role is to decide whether the query can be searched directly or should be decomposed into simpler sub-questions. If decomposition is necessary, the agent identifies logical break points and generates a set of targeted sub-queries. For instance, a complex query like ``How did Novo Nordisk's profit change between 2022 and 2023?'' would be decomposed into: \textit{1)}~``What was Novo Nordisk's profit in 2022?''~\textit{2)} ``What was Novo Nordisk's profit in 2023?'', and \textit{3)} ``What is the change in profit between 2022 and 2023?''. This decomposition allows for more focused subsequent searches and enables the system to handle comparative and multi-hop questions that require reasoning across multiple facts, aligning with plan $\rightarrow$ act patterns~\cite{erdogan2025plan,wang-etal-2023-plan,yao2023react}. If the original query is already simple enough, it is passed through directly. The output of the Planner agent is a set of one or more search queries $\{q_i\}$ to be processed.

%==================================
\subsubsection{Search Agent}
\label{sec:search_agent}
%==================================
The \textit{Search agent} executes sub-queries $q_i$ from the Planner agent in parallel (up to 12 concurrent searches). It employs a hybrid search strategy that combines keyword-based (sparse) and semantic (dense) retrieval, augmented by structured metadata filters to narrow the search space before chunk retrieval. For each query, the agent derives: \textit{1)} \textbf{a} semantic query for vector similarity search, \textit{2)} keywords for FTS, and \textit{3)} metadata filters (e.g., company names, dates, report types). This is the main mechanism by which metadata improves retrieval: it programmatically excludes irrelevant reports before dense and sparse ranking are applied at the chunk level.

\begin{figure*}[!hbt]
    \centering
    \includegraphics[width=0.8\textwidth]{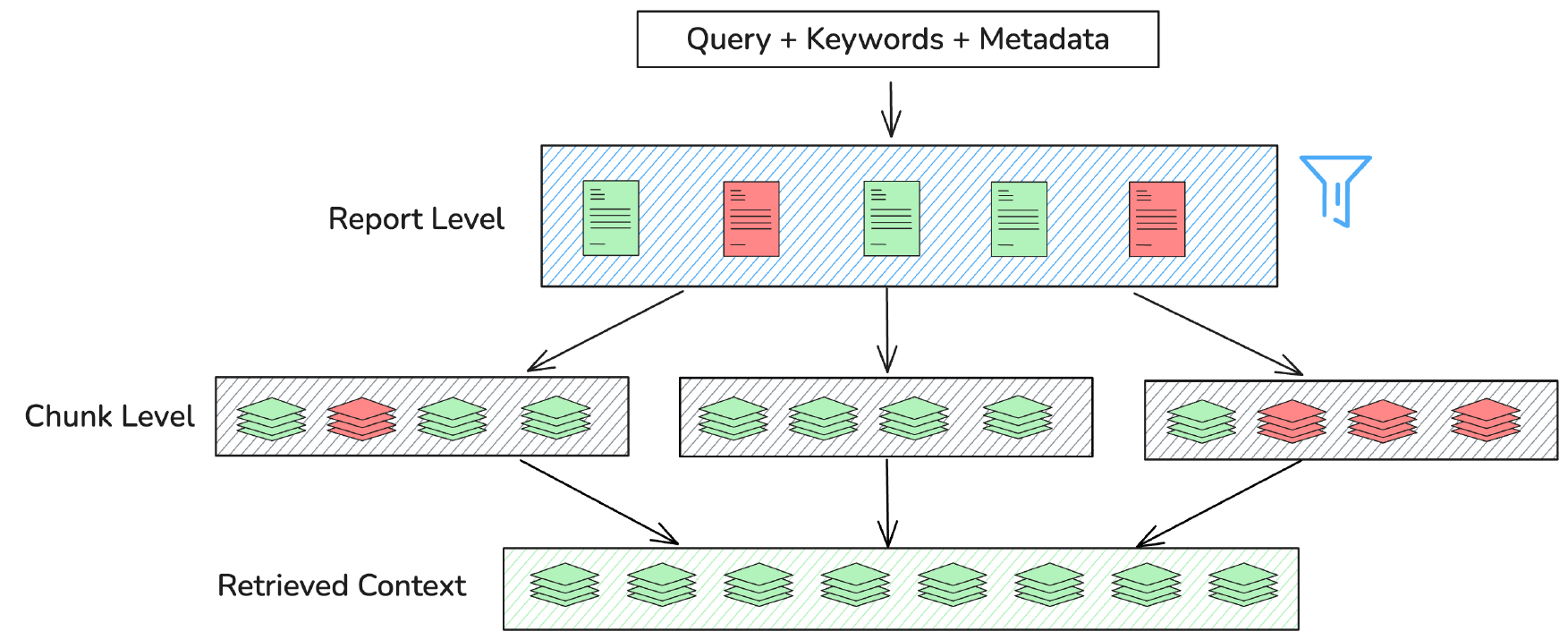}
    \caption{\textbf{Hybrid Search Process}: A query is first augmented with keywords and metadata filters. Relevant information is retrieved and used as pre-filters at the chunk level. A hybrid search is done over all relevant chunks and restricted to top-K}
    \label{fig:vector_search}
\end{figure*}

The available options for these metadata filters are dynamically populated at runtime by querying the vector database (cached with a one hour Time-To-Live (TTL) and refreshed upon new data additions), ensuring that agent-generated filter values are validated against the most current and valid set of choices. Agent-generated filters are validated programmatically against cached metadata using Pydantic~\cite{colvin_2025_15174950} models integrated with Pydantic AI\footnote{\url{https://github.com/pydantic/pydantic-ai}}; validation failures trigger LLM-correction and retry cycles. The hierarchical search process (Figure~\ref{fig:vector_search}) first applies these metadata filters to identify relevant reports. Chunks within these reports are then subjected to a hybrid search: dense retrieval (k-NN on vector embeddings) and sparse retrieval (FTS in LanceDB). Relevance scores are combined using Reciprocal Rank Fusion (RRF) to produce a top-K ranked list of chunks.

%==================================
\subsubsection{Validator Agent}
%==================================
The top-ranked chunks retrieved from a search are passed through the \textit{Validator agent}, which determines if a chunk could contribute to answering the user's question. We take advantage of parallel calls and process all chunks in one go. If no valid chunks are found, the system retries with reformulated queries (up to three attempts). If all attempts fail, the user is informed that no relevant content was found. We formalize the entire search pipeline in Algorithm~\ref{alg:full_pipeline}, and provide a sequence diagram in Figure~\ref{fig:sequence_digram}.

In Algorithm~\ref{alg:full_pipeline}, $Q$ denotes the original user query, $S$ the set of sub-queries, $m_s$ the metadata filters for sub-query $s$, $k_s$ the lexical query terms, $H_s$ the ranked search results, $V_s$ the validated chunks, and $A$ the accumulated validated set.

\begin{algorithm}[!hbt]
\caption{End-to-End Planning $\rightarrow$ Hybrid Retrieval $\rightarrow$ Validation Pipeline}
\label{alg:full_pipeline}
\begin{algorithmic}[1]
\Require User query $Q$
\Ensure Set $A$ of validated chunks or \texttt{NO\_ANSWER}

\State $S \gets \text{Planner}(Q)$ \Comment{Generate initial sub-questions}
\State $A \gets \emptyset$ \Comment{All validated chunks}
\State $\textit{round} \gets 1$

\While{$\textit{round} \leq \textit{MaxRounds}$}
    \ForAll{$s \in S$ \textbf{(in parallel, up to 20)}}
        \State $(k_s, m_s) \gets \text{PrepareSearch}(s)$ \Comment{Infer keywords and metadata filters}
        \State $H_s \gets \text{Search}(s, k_s, m_s)$ \Comment{Metadata pre-filtering + hybrid retrieval + RRF}
        \State $V_s \gets \{ h \in H_s \mid \text{Validate}(Q, s, h) = \textbf{true} \}$
        \State \textbf{emit} $V_s$ \Comment{Thread-local output}
    \EndFor
    \State $R \gets \bigcup_{s \in S} V_s$
    \State $A \gets A \cup R$
    \If{$|A| \geq 1$}
        \State \Return $A$
    \EndIf
    \If{$\textit{round} = \textit{MaxRounds}$}
        \State \textbf{break}
    \EndIf
    \State $S \gets \text{Replan}(Q, S)$ \Comment{Generate new sub-questions}
    \State $\textit{round} \gets \textit{round} + 1$
\EndWhile
\State \Return \texttt{NO\_ANSWER} \textbf{if} $A = \emptyset$ \textbf{else} $A$
\end{algorithmic}
\end{algorithm}
%

%%%%%%%%%%%%%%%%%%%%%%%%%%%%%%%%%%%%%%%%%%%%%%%%%%
\subsection{Post-Retrieval and Answer Generation}
%%%%%%%%%%%%%%%%%%%%%%%%%%%%%%%%%%%%%%%%%%%%%%%%%%
Post-retrieval organizes retrieved chunks and generates a final response grounded in the original documents.

%%%%%%%%%%%%%%%%%%%%%%%%%%%%%%%%%%%%%%%%%%%%%%%%%%
\subsubsection{Structured Input Construction}
\label{sec:structured_input}
\begin{figure*}[!hbt]
    \centering
   \includegraphics[width=0.7\textwidth]{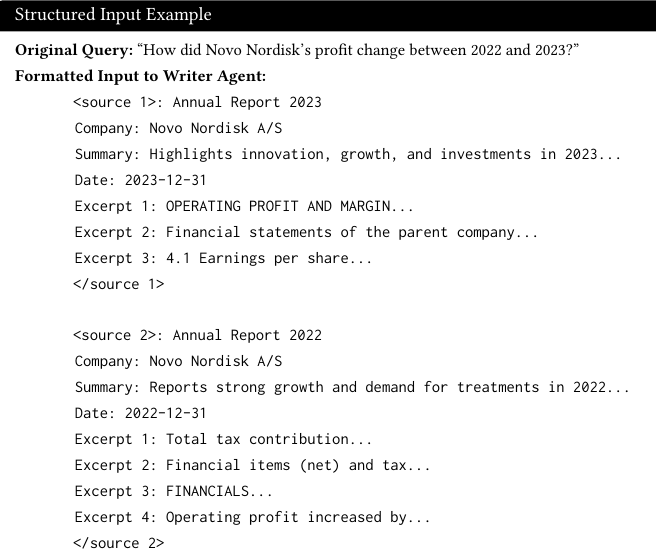}
    \caption{Sample of an input format}
    \label{fig:input_prompt}
\end{figure*}

%%%%%%%%%%%%%%%%%%%%%%%%%%%%%%%%%%%%%%%%%%%%%%%%%%
Once the validated chunks are retrieved, they are grouped and formatted into a structured input that serves as the context for the \textit{Writer Agent}. This structured format is designed to improve both grounding and interpretability, inspired by a prompt format\footnote{https://blog.skypilot.co/deepseek-rag/} that also aligns with findings from~\cite{tan2024structuredInput,merrick2024arctic}. Chunks are grouped by their originating report (report\_id), and within each group, they are sorted by relevance and labeled as individual excerpts. This grouping ensures that the Writer agent can reason over coherent sets of information rather than isolated fragments. The final input to the Writer agent is a concatenation of these structured report sections, each enclosed in a \texttt{<source>} tag. An example of the format is represented in Figure~\ref{fig:input_prompt}. Users do not create this prompt structure manually; they submit natural-language questions, while the system constructs the internal retrieval and generation prompts automatically.
%

% %

\subsubsection{Writer Agent}

The \textit{Writer Agent} synthesizes its final response from four inputs: the original user query; grouped and validated chunks; document metadata (for source info); and a precision-and-citation prompt template. Answers include inline citations and a programmatically appended source list (refer to Figure~\ref{fig:iter1_output_example_comparison}).

%=====================================
\subsection{Agent Communication}
%=====================================
MimirRAG's inter-agent communication leverages Pydantic AI, a lightweight framework ensuring asynchronous operations and type-safe, structured interactions. Each agent is a self-contained class with an asynchronous \texttt{run} method, configured with task-specific prompts, tools (e.g., hybrid search), and Pydantic models for its inputs and outputs. While the overall workflow proceeds through logically sequential stages, the system utilizes asynchronous execution to achieve parallel processing, particularly for performance-critical search and validation phases.

\begin{figure*}[!hbt]
    \centering
    \includegraphics[width=.8\textwidth]{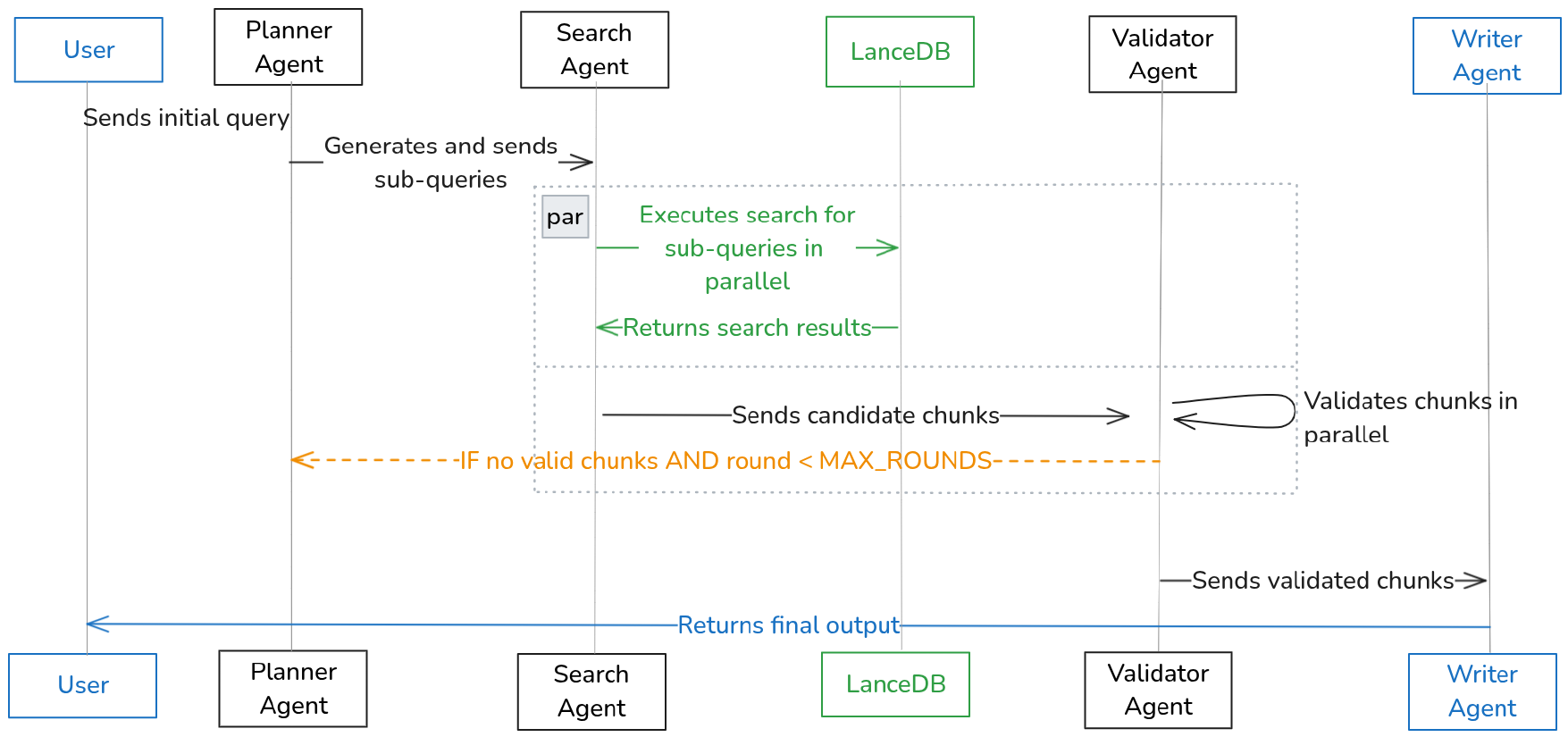}
    \caption{Sequence diagram illustrates the interaction between agents in the hierarchical retrieval and validation pipeline. Green arrows denote the parallel search against the vector store, black arrows show data flow between agents, the dashed orange arrow represents the fallback to replanning when no chunk is validated and \texttt{round}\(<\)\texttt{MAX\_ROUNDS}, and the blue line denote the final response back to the user.}
    \label{fig:sequence_digram}
\end{figure*}

Figure~\ref{fig:sequence_digram} illustrates this \textit{interaction logic}, showcasing how agents exchange structured messages. This architecture supports concurrent execution (parallel searches and validations) and Pydantic AI orchestration logic that handles fallbacks and retries. Data integrity is paramount, thus all agent outputs are validated against their Pydantic models. This strict validation, coupled with automatic retries (three by default) on failure, ensures reliability, simplifies error handling, and aids system maintenance.

%%%%%%%%%%%%%%%%%%%%%%%%%%%%%%%%%%%%%%%%%%%%%%%%%%
\subsection{Conversation Agent}
\label{sec:conversation_agent}
%%%%%%%%%%%%%%%%%%%%%%%%%%%%%%%%%%%%%%%%%%%%%%%%%%
While the five agents constitute the core RAG workflow for processing and answering queries from scratch, MimirRAG incorporates an additional \textit{Conversation agent} that acts as an intelligent routing layer, managed by the \texttt{ConversationService} on the backend. For the Conversation agent, the RAG workflow is a tool that is used to answer the user's question. The primary role of the Conversation agent is to manage the dialogue flow between the user and the RAG workflow, particularly for follow-up questions within an ongoing conversation. When a user submits a message, the \texttt{ConversationService} first determines if it's the initial query of a conversation or a subsequent one. For the first message, the query is directly routed to the full RAG workflow, starting with the Planner agent, to retrieve and synthesize an answer. For subsequent messages, the Conversation agent is invoked. It analyzes the new user message in the context of the preceding conversation history. Based on this analysis, and guided by its specific prompt, it decides whether the new message requires a full, new retrieval process (e.g., the user asks a completely new question) or if it can be directly answered based on already retrieved content.

This approach makes user interaction more natural and efficient by avoiding redundant RAG workflow executions for conversational turns that do not necessitate new information retrieval. The Conversation agent thus enhances the user experience by providing more context-aware and responsive interactions. This also makes the system more modular and easier to extend with new features, as they can simply be added as new tools to the Conversation agent.

%%%%%%%%%%%%%%%%%%%%%%%%%%%%%%%%%%%%%%%%%%%%%%%%%%
\subsection{Prompt Engineering}
\label{subsec:prompt_engineering}
\begin{figure}[!hbt]
    \centering
   \includegraphics[width=\columnwidth]{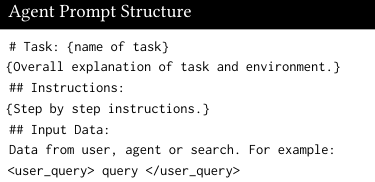}
    \caption{Sample of an agent prompt structure}
    \label{fig:agent_prompt}
\end{figure}

%%%%%%%%%%%%%%%%%%%%%%%%%%%%%%%%%%%%%%%%%%%%%%%%%%
Each agent uses a task-specific prompt. Prompt templates remain static across user sessions, while the content is dynamically filled based on the query- and chunk-specific content. Users interact with the system through a standard chat/query interface and provide only the natural-language question. They do not generate the internal prompts shown in Figure~\ref{fig:agent_prompt}. By using prompt templates, we ensure consistency across the RAG workflow and make it easier to fine-tune individual stages independently. In the initial development phase, our priority was to establish a functional end-to-end workflow for each agent. Consequently, we focused on composing clear and straightforward system- and user-prompts that accurately described each agent's intended task, enabling the required function calls and basic output formatting. While no systematic A/B testing was conducted, prompts were iteratively refined on an ad-hoc basis in response to observed errors or unexpected behavior during development and user interaction. For instance, the Search agent's prompt was updated to default to searching within the last 12 months if no specific date was provided by the user, a change driven by observing common user interaction patterns.

Our approach to prompt design aims to enhance agent reliability and performance. It is guided by~\cite{Anthropic_2024,Zhang_HowWeBuildEffectiveAgents_2025}: ensuring the agent is aware of its operational environment and available tools, clearly documenting tool functionalities from the agent's perspective, and designing instructions by considering what information and guidance an agent would need to perform its task effectively. The overall prompt template structure used is shown in Figure~\ref{fig:agent_prompt}.
%
%================================================
\subsection{Benchmarking}
\label{sec:experimental_setup}
%================================================
We employ FinanceBench, a curated dataset tailored for assessing financial question-answering systems. It comprises over 10,000 question-answer pairs derived from 361 public filings released between 2015 and 2023, specifically designed to benchmark RAG systems and LLMs in financial contexts. Each entry includes a question, a gold-standard answer, and annotated evidence text, grounding the response in a specific location within a filing. This structure makes FinanceBench ideal for evaluating both retrieval accuracy and reasoning capabilities. FinanceBench tests two core competencies. They are \textit{1)} Retrieval: The ability to locate relevant evidence within a realistic corpus of SEC filings and \textit{2)} Reasoning: The capacity to perform logical and numerical reasoning over retrieved evidence to address analyst-style questions. The dataset includes fact-based, reasoning-based, and context-aware questions, mirroring typical financial analyst research questions. 

FinanceBench features three primary question categories.~\textit{1)}~Domain-relevant: A fixed set of 25 common analyst checks, e.g., \textit{Has Company X paid a dividend in fiscal year Y?}~\textit{2)}~Novel-generated: Analyst-crafted questions simulating realistic variations of typical Q\&A.~\textit{3)}~Metric-generated:~Programmatically generated from 18 baseline items in financial statements, including derivatives. Subcategories based on reasoning types, numerical reasoning, information extraction, and logical reasoning, enable detailed performance analysis.

\subsubsection{Dataset Subset}
We use the 150-question open-source evaluation subset released by the authors for their original experiments, as the full FinanceBench dataset is not publicly available. This subset comprises: 50 domain-relevant questions, 50 novel analyst-generated questions, and 50 metric-based questions derived from financial statements, covering 32 companies. This combination ensures a balanced assessment of direct information retrieval and nuanced reasoning tasks.

\subsubsection{Answer Checking}
The benchmark questions are evaluated against the gold-standard answers which is performed by a Judge Agent, an LLM-based evaluator (specifically, GPT-4o in our setup). The Judge Agent categorizes each answer into three labels. \textit{1)~Correct} means the answer matches the gold label and is grounded in the correct evidence. Minor discrepancies (e.g., small rounding errors) are tolerated; \textit{2)~Incorrect} means the answer is factually wrong, contains calculation errors, or contradicts the gold answer;~\textit{3)~Failure to Answer} The model explicitly indicates it cannot answer the question. This aligns with the evaluation criteria used in the original FinanceBench paper, where they used human evaluators to score the answers. In addition to the label, the Judge Agent is also prompted to provide an explanation of the label.

\subsubsection{Retrieval}
In addition to evaluating the final generated answer, we also assess the performance of the retrieval component using several metrics:

\begin{itemize}
    \item \textbf{Hit Rate (Hit@k):} This metric measures the percentage of questions for which the correct \textit{document} containing the grounding evidence (as defined by FinanceBench) is found within the top $k$ retrieved documents. We specifically report Hit@1 and Hit@5, indicating whether the document containing the correct evidence was ranked as the first or within the top five retrieved documents, respectively. This assesses the retrieval system's ability to identify and rank the most relevant source documents.
    \item \textbf{Chunk Similarity:} To evaluate the quality of the retrieved chunks at a content level, for each question, we identify the single retrieved chunk (from the top-k set) that has the highest similarity to the gold-standard evidence. These individual maximum similarity scores are then averaged across all 150 questions in the benchmark dataset. We provide two different similarity metrics:
    \begin{itemize}
        \item \textit{Cosine Similarity:} Calculated on the text embeddings (`snowflake-arctic-embed-m-v2.0'), this measures the semantic similarity between the retrieved chunk and the gold evidence. Higher values (close to one) indicate greater semantic relevance.
        \item \textit{Levenshtein Similarity:} This metric quantifies the textual similarity based on the minimum number of single-character edits (insertions, deletions, or substitutions) required to change one text into the other. We report a normalized similarity score where higher values indicate greater textual overlap with the gold evidence. We remove all markdown formatting from the retrieved chunks before calculating the similarity, as the evidence text is formatted as plain text.
    \end{itemize}
\end{itemize}
These retrieval metrics provide insights into the effectiveness of the system's ability to locate and rank the precise information needed to answer the user's query, complementing the end-to-end evaluation performed by the Judge Agent.

\subsubsection{Existing Benchmarks and Retrieval Strategies}
Table~\ref{tab:bench_settings} defines the retrieval strategies evaluated on FinanceBench, from the original paper and other works. Crucially, to establish an updated oracle baseline with one of our evaluated LLMs, we ran the oracle evidence pages (perfect retrieval) with GPT-4.1 as the generator. This configuration achieved 93.3\% accuracy using our Judge Agent evaluation, providing an updated reference point for what is achievable with perfect context and current models.
\begin{table*}[!hbt]
\centering
\caption{Retrieval strategies evaluated on FinanceBench.}
\label{tab:bench_settings}
\begin{adjustbox}{width=.8\textwidth}
\begin{tabular}{|c|l|}
\hline
\textbf{Retrieval Strategy}     & \multicolumn{1}{c|}{\textbf{Description}}                                                                                                              \\ \hline
\textbf{Closed book}            & Model relies solely on pre-trained knowledge, without external context or documents.                                                                   \\ \hline
\textbf{Shared vector store}    & All filings indexed in a single vector database; retrieves top-k passages across all documents.                                                        \\ \hline
\textbf{Single vector store}    & Each filing has its own vector index; retrieval is scoped to the relevant filing.                                                                      \\ \hline
\textbf{Full filing in context} & Entire filing text (up to the model's token limit) is provided as in-context input.                                                                    \\ \hline
\textbf{Oracle}                 & Supplies only the exact "gold-standard" evidence pages containing the answer.                                                                          \\ \hline
\textbf{Mafin 2.5}              & \textbf{\begin{tabular}[c]{@{}l@{}}Reasoning-based RAG via PageIndex: hierarchical tree navigation \\ over document structure; replaces vector similarity with \\ LLM-driven structured retrieval.\end{tabular} }
                                                      \\ \hline
\textbf{OODA loop}              & Iterative cycle: Observe, Orient, Decide, Act; applied to retrieval and reasoning.                                                                     \\ \hline
\end{tabular}
\end{adjustbox}
\end{table*}
Table~\ref{tab:financebench} presents performance metrics for various models and strategies on the 150-question subset.
\begin{table*}[!hbt]
\centering
\caption{Performance summary on the 150-question FinanceBench benchmark. Results marked with * are from~\cite{islam2023financebench}.}
\label{tab:financebench}
\begin{adjustbox}{width=.8\textwidth}
\begin{tabular}{|c|c|c|c|}
\hline
\textbf{Retrieval Strategy}          & \textbf{Context Width} & \textbf{Model}  & \textbf{Accuracy (\%)} \\ \hline
Closed book*                         & --                     & GPT-4-Turbo     & 9\%                    \\ \hline
Shared vector store*                 & --                     & GPT-4-Turbo     & 19\%                   \\ \hline
Single vector store per filing*      & --                     & GPT-4-Turbo     & 50\%                   \\ \hline
Full filing in context*              & 95K tokens             & GPT-4-Turbo     & 79\%                   \\ \hline
Oracle (evidence pages)*             & Evidence pages         & GPT-4-Turbo     & 85\%                   \\ \hline
\textbf{Our Oracle (evidence pages)} & Evidence pages         & GPT-4.1    & \textbf{93.3\%}        \\ \hline
Databricks                           & 64K tokens             & o1-preview-2024 & $\sim$75\%             \\ \hline
OODA                                 & --                     & GPT-3.5-turbo   & 82\%                   \\ \hline
Mafin 2.5 (Vectify AI)               & --                     & GPT-4o          & \textbf{98.7}\%        \\ \hline
\end{tabular}
\end{adjustbox}
\end{table*}

Databricks~\cite{Leng2024LongCRA} achieved $\sim$75\% accuracy using o1-preview with long-context RAG, leveraging a 64K-token context window to process top-125 filing chunks. OODA~\cite{nguyen2024enhancing} employed an Observe-Orient-Decide-Act (OODA) loop with retrieval, reaching 82\% accuracy with GPT-3.5-turbo, surpassing their fine-tuned retrieval approach. They removed 9 questions due to broken or missing document sources, deeming 141 of 150 questions suitable for benchmarking, though specific details about the disregarded questions are not provided. 

Mafin 2.5~\cite{vectify2025mafingithub}: Vectify AI's system achieves 98.7\% accuracy with GPT-4o, built on PageIndex, a reasoning-based RAG framework that replaces vector similarity with hierarchical tree navigation over document structure. PageIndex constructs a semantic tree from document headers and LLM-generated summaries, enabling structured reasoning during retrieval rather than embedding-based matching. An earlier version of the system documentation also described reinforcement learning with Monte Carlo Tree Search (MCTS) as part of the pipeline\footnote{Per the Mafin 2.5 GitHub repository prior to its October 2025 revision; current documentation describes the system solely as reasoning-based RAG powered by PageIndex.}, though the full implementation details have not been published. Mafin adjusts 12 of the 150 benchmark questions: 6 removed as benchmark errors, 5 corrected for multiple valid answers (MVA), and 1 for same evidence, different conclusion (SEDC). These adjustments are disclosed\footnote{\url{https://github.com/VectifyAI/Mafin2.5-FinanceBench/tree/main/human_evaluations}}. Under the original unadjusted gold standards, their accuracy would be approximately 90.7\% (136/150), close to our Oracle baseline. Given the limited methodological transparency, Mafin 2.5 is excluded from direct replicable comparison with MimirRAG. With the exception of~\cite{Leng2024LongCRA}, neither OODA nor Mafin 2.5 disclose prompt templates or the number of reasoning rounds used during benchmarking, limiting direct comparability. We find four key takeaways for RAG system development based on these previous results:
\begin{itemize}
    \item \textit{Value of Retrieval:} Basic RAG approaches demonstrate substantial gains over closed-book models. For instance, simply providing the full filing context boosted accuracy from 9\% (GPT-4-Turbo, closed-book) to 79\%, and long-context RAG systems like Databricks achieve $\sim$75\%. This underscores that access to relevant information is a primary driver of performance.
    \item \textit{Better reasoning:} Once retrieval is in place, employing agentic loops like OODA (82\% accuracy), significantly improves results, bringing performance closer to the 85\% oracle baseline. This suggests that refining \textit{how} information is identified, selected and reasoned about is crucial.
    \item \textit{Improved base model}: Our new oracle baseline (93.3\%), highlights that as LLMs become more capable, the performance ceiling for RAG systems, even with ideal retrieval, inherently rises.
    \item \textit{Beyond the Oracle with Advanced Architectures:} State-of-the-art pipelines like Mafin 2.5 (98.7\% accuracy) demonstrate that it's possible to surpass the traditional oracle. This is achieved not only through more capable LLMs (GPT-4o vs. GPT-4-Turbo in the original oracle) but also through structured, reasoning-based retrieval that preserves document hierarchy, rather than relying on flat embedding similarity. This implies that the retrieval ceiling can be raised by rethinking how documents are indexed and navigated, not solely by improving embedding or generation models.
\end{itemize}
We aim to assess MimirRAG's performance relative to replicable established benchmarks, such as the original FinanceBench paper's results and our own updated Oracle baseline. Our evaluation leverages the entire 150-question FinanceBench subset, and we will discuss potential inherent errors in the benchmark.

%%%%%%%%%%%%%%%%%%%%%%%%%%%%%%%%%%%%%%%%%%%%%%%%%%
\section{Experimental Analysis}
%%%%%%%%%%%%%%%%%%%%%%%%%%%%%%%%%%%%%%%%%%%%%%%%%%
Benchmarking provides crucial insights into a system's technical performance under controlled conditions.

\subsection{Selected LLM Variants}
We evaluated our pipeline using OpenAI's GPT-4 language model available at the partner company, chosen to explore trade-offs between model size, response quality, and application programming interface (API) cost. GPT-4.1 was our upper-bound baseline for language generation and instruction following. GPT-4.1 Mini is a distilled variant offering a balance of quality and cost, while GPT-o4 Mini is a reasoning-optimized variant. Table~\ref{tab:llm-comparison-exp} details these models where MMLU score refers to a model's performance on the massive multitask language understanding (MMLU) benchmark. We calculated the relative cost per \% accuracy of each model based on the cost of the input and output tokens divided by the accuracy of the model\footnote{It does not account for potential differences in token usage between models.}
\[
\text{Relative Cost} = \frac{\text{Input Price} + \text{Output Price}}{\text{Accuracy}}
\]

\begin{table*}[!hbt]
\centering
\caption{Comparison of LLMs used in experiments.}
\label{tab:llm-comparison-exp}
\begin{adjustbox}{width=.8\textwidth}
\begin{tabular}{|c|c|c|c|c|}
\hline
\textbf{Model} & \textbf{Context Length} & \textbf{MMLU Score} & \textbf{Pricing (USD per 1M tokens)} & \textbf{Release} \\ \hline
GPT-4o         & 128K tokens             & 87.0\%              & \$2.50 (input) / \$10.00 (output)        & May 2024         \\ \hline
GPT-4.1        & 1M tokens               & 90.2\%              & \$2.00 (input) / \$8.00 (output)         & Apr 2025         \\ \hline
GPT-4.1-mini   & 1M tokens               & 87.5\%              & \$0.40 (input) / \$1.60 (output)         & Apr 2025         \\ \hline
GPT-o4-mini    & 200K tokens             & ?\%                 & \$1.10 (input) / \$4.40 (output)         & Apr 2025         \\ \hline
\end{tabular}
\end{adjustbox}
%\vspace{1ex}
\parbox{0.9\linewidth}{
\small \textit{Note: Pricing information from OpenAI API documentation is subject to change.}
}
\end{table*}

\subsection{Environment Setup}
The system is deployed using Azure Container Apps, with the API running as the primary container and both the UI and proxy deployed as sidecars within the same app. The containers use different base images: Debian 12 (Slim Bookworm) for the API and UI, and Alpine for the proxy. Resource allocations are defined in Azure as follows: API: 0.75 vCPU and 2.6 GB RAM; UI: 0.5 vCPU and 0.7 GB RAM and Proxy: 0.25 vCPU and 0.5 GB RAM. A NVIDIA Tesla T4 (with 2,560 CUDA cores, 320 tensor cores and 16 GB of GDDR6 GPU memory) is provisioned on demand when GPU acceleration is required. For the first iteration, all experiments are run with max rounds set to three and max retrieved reports set to five. Concurrency is adjusted based on model and rate limits (Algorithm~\ref{alg:full_pipeline}). Temperature for all agents is set to 0.1, except for o4-mini, which at the time of testing did not accept a temperature setting.

%%%%%%%%%%%%%%%%%%%%%%%%%%%%%%%%%%%%%%%%%%%%%%%%%%
\subsection{Prototyping}
%%%%%%%%%%%%%%%%%%%%%%%%%%%%%%%%%%%%%%%%%%%%%%%%%%
%
\begin{figure*}[!hbt]
    \centering
    \includegraphics[width=0.8\textwidth]{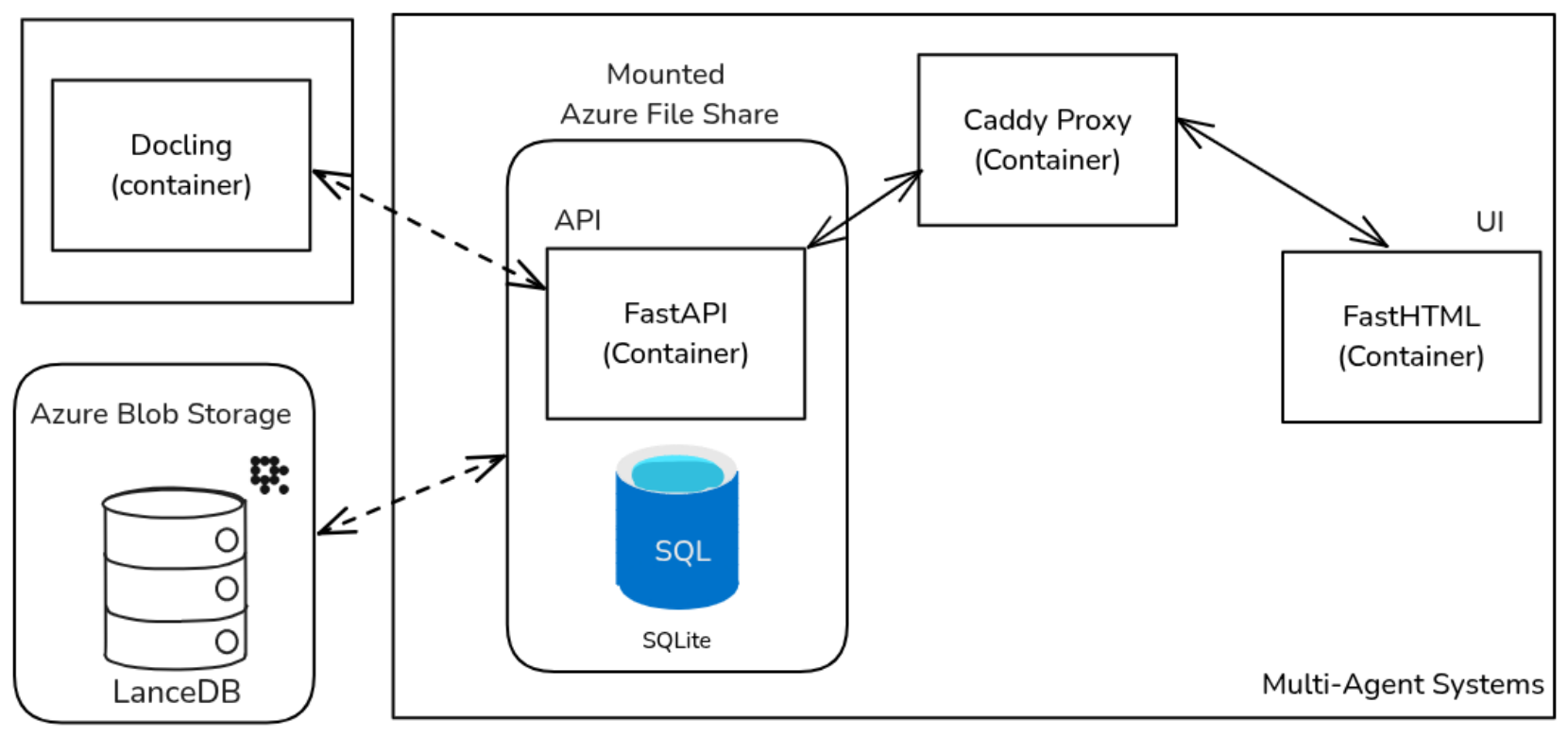}
    \caption{High-level Azure deployment diagram for MimirRAG.}
    \label{fig:azure_arch}
\end{figure*}
\begin{figure*}[!hbt]
    \centering
    \includegraphics[width=.8\textwidth]{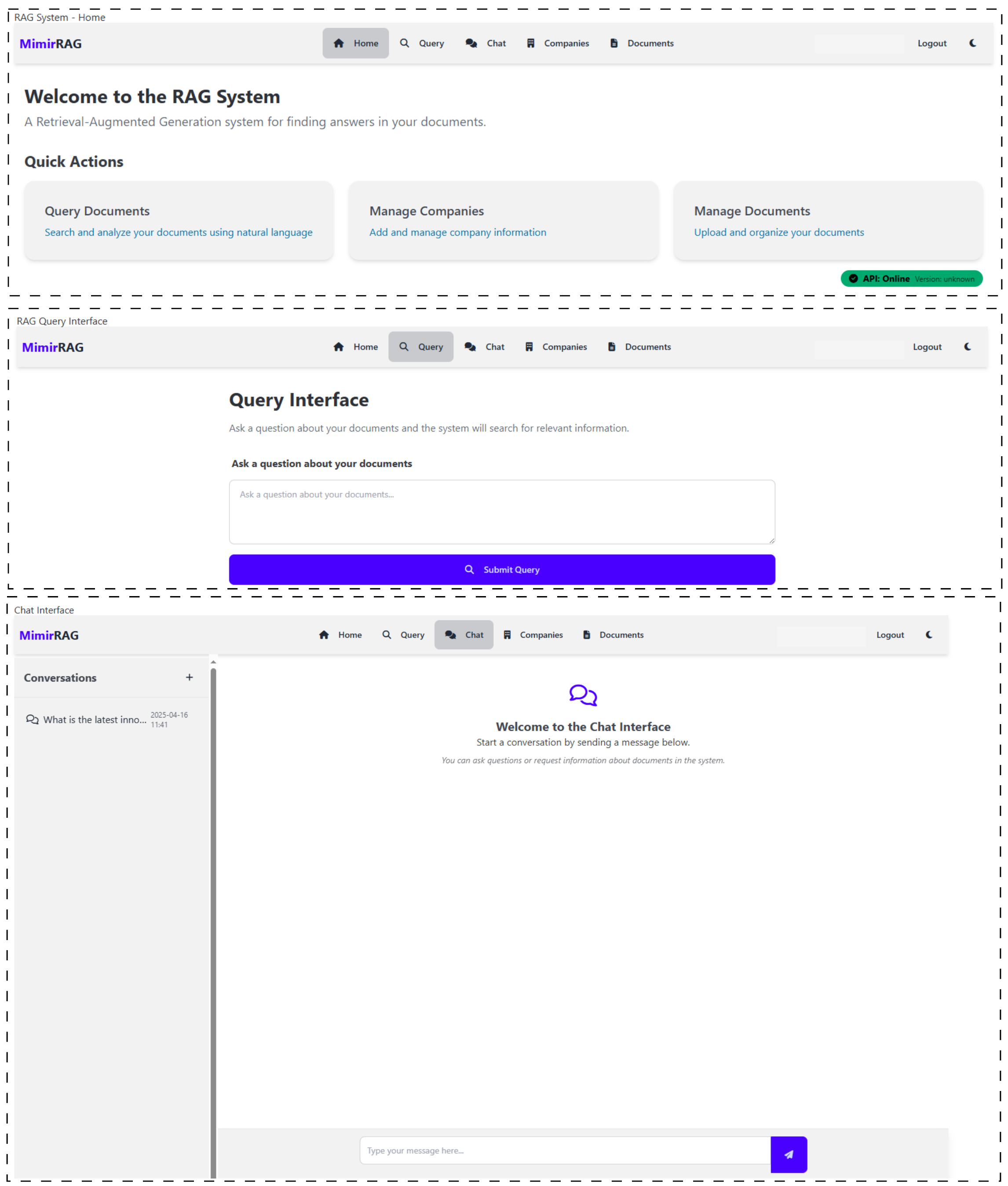}
    \caption{Screenshots of MimirRAG prototype: home page (top), query interface (middle) and chat interface (bottom)}
    \label{fig:ui_chat1}
\end{figure*}

MimirRAG was prototyped as a web application deployed on Azure Container Apps (see Figure~\ref{fig:azure_arch}). The backend, built with FastAPI\footnote{\url{https://fastapi.tiangolo.com/}} for asynchronous processing, interfaces with LanceDB (in Azure Blob Storage) for vector/metadata storage and SQLite (on Azure Files) for relational data. The UI enabled users to submit natural-language queries, manage conversational history, and dynamically update the knowledge base by uploading documents and managing company profiles. Internal prompts for the agents were predefined by the system and not manually authored by end-users. Evaluation primarily focused on the chat interface (refer to bottom of Figure~\ref{fig:ui_chat1}).

Figure~\ref{fig:ui_chat1} displays three of the five primary webpages of MimirRAG (the remaining two are omitted due to space constraints). The top section illustrates the landing page, the middle section shows the query interface, and the bottom section presents the main chat interface.

\subsection{Ablation Study}
The study establishes a baseline using Naive RAG (with a shared vector store) and a custom chunking and prompt setup, ensuring it reflects updated embedding and LLM models. It then performs ablations to assess the effects of various LLM variants and retrieval configurations (see Table~\ref{tab:ite_1}). We focused our ablation studies on the GPT-4.1-mini model, as it was the cheapest model, allowing us to test many configurations. The first iteration focused on establishing a baseline of our system and identifying initial performance characteristics. The results from these configurations informed the design of iteration two.
\begin{table*}[!hbt]
\centering
\caption{Ablation settings for iteration 1}
\label{tab:ite_1}
\begin{adjustbox}{width=.8\textwidth}
\begin{tabular}{|l|l|}
\hline
\multicolumn{1}{|c|}{\textbf{Configuration}} & \multicolumn{1}{c|}{\textbf{Description}}                                                                                                               \\ \hline
Naive RAG                                    & \begin{tabular}[c]{@{}l@{}}Re-implementing the original Shared Vector \\ Store RAG system from the FinanceBench paper.\end{tabular}                      \\ \hline
Retrieval agents impact                      & \begin{tabular}[c]{@{}l@{}}Testing the effect of removing the Planner, Search, \\ or Validator agent individually from the MimirRAG setup.\end{tabular} \\ \hline
Metadata filtering                           & \begin{tabular}[c]{@{}l@{}}Comparing performance with and without \\ metadata constraints during search.\end{tabular}                                   \\ \hline
LLM variant                                  & Evaluating GPT-4.1, GPT-4.1 Mini, and GPT-o4 Mini.                                                                                                      \\ \hline
No. of Chunks (Search Breadth)               & \begin{tabular}[c]{@{}l@{}}Varying the number of retrieved chunks per \\ sub-question (k=10 vs. k=30).\end{tabular}                                     \\ \hline
\end{tabular}
\end{adjustbox}
\end{table*}
These settings constitute the main ablation study for the paper (refer Table~\ref{tab:ite_1}). They isolate the impact of metadata filtering, individual retrieval-stage agents, search breadth, and LLM choice on the end-to-end system.
%================================================
\subsection{First Iteration}
\label{sec:iter1}
%================================================

\subsubsection{Performance of Models}
\label{sec:iter1-results}

Table~\ref{tab:model-performance-iter1} summarizes the performance across different models and retrieval configurations in Iteration 1. Our reimplemented Naive RAG setup, using GPT-4.1-mini with our enhanced preprocessing (Docling for PDF to Markdown conversion and Markdown-aware chunking) and snowflake-arctic-embed-m-v2.0 embeddings, achieved 65.3\% accuracy. This significantly outperforms the 19\% baseline reported in the original FinanceBench paper for their GPT-4-Turbo Naive RAG configuration. This substantial improvement underscores the collective impact of more advanced LLMs, better embedding models, and a more robust document preprocessing and chunking pipeline than what was likely employed in the original study. Importantly, we do this with a small top-k of 10, compared to previously discussed results by Databricks using top-128 ($\sim 75\%$).

\begin{figure*}[!hbt]
    \centering
    \begin{adjustbox}{width=0.8\textwidth}
    \includegraphics{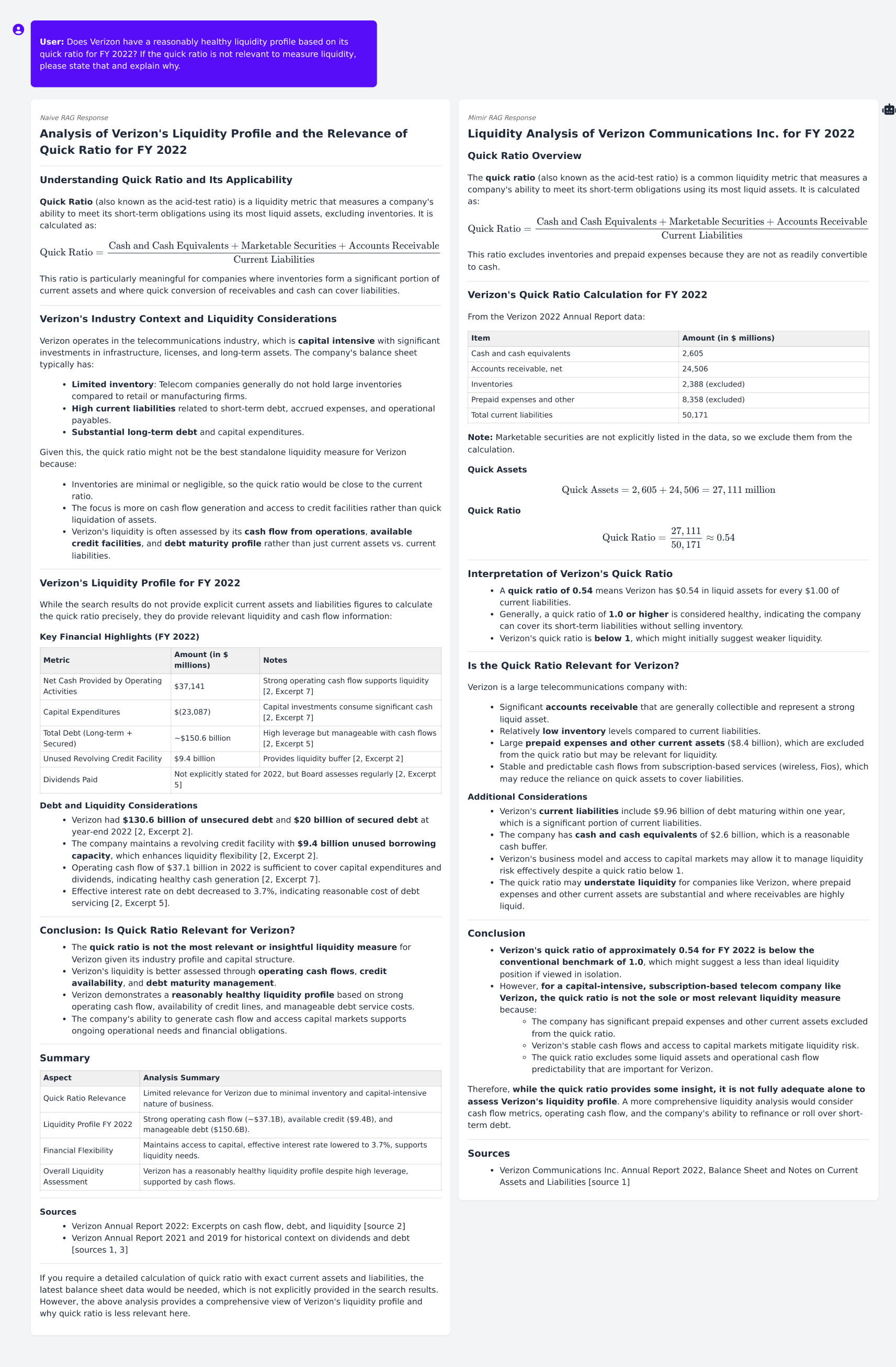}
    \end{adjustbox}
    \caption{Output example comparing Naive RAG (left) with MimirRAG (right, using GPT-4.1-mini) for a question on Verizon's FY 2022 Quick Ratio. Naive RAG fails to retrieve the correct data from multiple documents, while MimirRAG successfully identifies the correct document and computes the ratio.}
    \label{fig:iter1_output_example_comparison}
\end{figure*}

The full MimirRAG system, in its best Iteration 1 configuration (GPT-4.1, Mimir, 10), achieved 82.7\% accuracy. This result approached the original FinanceBench Oracle performance (85\% with GPT-4-Turbo) and demonstrates strong capability even before the targeted optimizations introduced in Iteration 2. In terms of relative cost, the full MimirRAG configuration with GPT-4.1-mini (Mimir, 10) was the most cost-effective among the MimirRAG setups, with a calculated relative cost of \$0.026 per \% accuracy. The full GPT-4.1 model in the same configuration was approximately 4.65 times more expensive per percentage point of accuracy.
\begin{table*}[!hbt]
\centering
\caption{Iteration 1: Performance comparison of LLMs and configurations. Results marked with * are from~\cite{islam2023financebench}. ** GPT-4.1-mini is used as the Search and Validator Agent in this setting for improved latency (this model is cheaper but not accounted for in the cost column).}
\label{tab:model-performance-iter1}
\begin{adjustbox}{width=.8\textwidth}
\begin{tabular}{|c|c|c|c|c|c|}
\hline
\textbf{Model} & \textbf{Configuration} & \textbf{Correct}      & \textbf{Incorrect}   & \textbf{Failed to answer} & \textbf{Rel. Cost (\$/\% Acc.)} \\ \hline
GPT-4-turbo*   & Naive RAG*             & 29 (19\%)             & 20 (13\%)            & 101 (68\%)                & N/A                             \\ \hline
GPT-4.1-mini   & Oracle                 & 137 (91.3\%)          & 13 (8.7\%)           & 0 (0\%)                   & N/A                             \\ \hline
GPT-4.1        & Oracle                 & 140 (93.3\%)          & 10 (6.7\%)           & 0 (0\%)                   & N/A
                        \\ \hline
GPT-4.1-mini   & 10, No Meta            & 93 (62.0\%)           & 49 (32.7\%)          & 8 (5.3\%)                 & 0.032                           \\ \hline
GPT-4.1-mini   & 10, Naive RAG          & 98 (65.3\%)           & 43 (28.6\%)          & 9 (6\%)                   & 0.031                           \\ \hline
GPT-4.1-mini   & 10, Naive RAG, Planner & 104 (69.3\%)          & 41 (27.3\%)          & 5 (3.3\%)                 & 0.029                           \\ \hline
GPT-4.1-mini   & 10, No Val             & 107 (71.3\%)          & 42 (28.0\%)          & \textbf{1 (0.7\%)}        & 0.028                           \\ \hline
GPT-4.1-mini   & Mimir, 10                     & 114 (76.0\%)          & 34 (22.7\%)          & 2 (1.3\%)                 & \textbf{0.026}                  \\ \hline
GPT-4.1-mini   & Mimir, 30                     & 114 (76.0\%)          & 31 (20.7\%)          & 5 (3.3\%)                 & \textbf{0.026}                  \\ \hline
GPT-4.1        & 10, Naive RAG, Planner & 118 (78.7\%)          & 31 (20.7\%)          & \textbf{1 (0.7\%)}        & 0.127                           \\ \hline
o4-mini**      & Mimir, 10                     & 119 (79.3\%)          & 28 (18.7\%)          & 3 (2.0\%)                 & 0.069                           \\ \hline
GPT-4.1        & Mimir, 10                     & \textbf{124 (82.7\%)} & \textbf{22 (14.7\%)} & 4 (2.7\%)                 & 0.121                           \\ \hline
\end{tabular}
\end{adjustbox}
\end{table*}
The qualitative difference between the Naive RAG approach and the MimirRAG system is illustrated in Figure~\ref{fig:iter1_output_example_comparison}. It presents an example where the Naive RAG system (left side of Figure~\ref{fig:iter1_output_example_comparison}) struggles with document disambiguation for a question about Verizon's Quick Ratio, retrieving information from three different documents and failing to find the necessary data. In contrast, the MimirRAG system (right side of Figure~\ref{fig:iter1_output_example_comparison}), even with the cost-effective GPT-4.1-mini model, correctly identifies the relevant Verizon annual report, extracts the necessary figures, and accurately calculates the Quick Ratio. This highlights MimirRAG's improved ability to precisely locate and utilize correct information from a larger corpus, a key advantage of its structured agentic approach and metadata utilization over simpler RAG techniques.
\subsubsection{Retrieval Analysis}
Table~\ref{tab:doc-performance-iter1} presents document-level retrieval metrics. Metadata filtering showed a clear advantage. For instance, GPT-4.1-mini (Mimir, 10) achieved Hit@1 of 0.65 and Hit@5 of 0.81 compared to 0.53 and 0.73 for the No Meta configuration.
Compared to the Naive RAG configuration, we observed large gains in Hit@1 while the gap is smaller at Hit@5. Metadata also led to a large reduction in the average number of documents retrieved, and the Validator agent effectively reduced the number of chunks retrieved. This supports the proposed mechanism that metadata improves answer quality mainly by restricting the candidate document set before chunk-level retrieval, thereby reducing ambiguity and irrelevant evidence.

\begin{table*}[!hbt]
\centering
\caption{Iteration 1: Retrieval performance across configurations.}
\label{tab:doc-performance-iter1}
\begin{adjustbox}{width=.8\textwidth}
\begin{tabular}{|c|c|c|c|c|c|}
  \hline
  \multicolumn{1}{|c|}{\textbf{Model}}
    & \multicolumn{1}{c|}{\textbf{Setting}}
    & \multicolumn{1}{c|}{\textbf{Hit@1}}
    & \multicolumn{1}{c|}{\textbf{Hit@5}}
    & \multicolumn{1}{c|}{\textbf{Avg docs}}
    & \textbf{Avg chunks} \\ \hline

  \multicolumn{1}{|c|}{GPT-4.1}
    & \multicolumn{1}{c|}{Mimir, 10}
    & \multicolumn{1}{c|}{0.67}
    & \multicolumn{1}{c|}{0.87}
    & \multicolumn{1}{c|}{2.02}
    & 8.19                \\ \hline

  \multicolumn{1}{|c|}{GPT-4.1}
    & \multicolumn{1}{c|}{10, Naive RAG, Planner}
    & \multicolumn{1}{c|}{0.47}
    & \multicolumn{1}{c|}{0.88}
    & \multicolumn{1}{c|}{14.23}
    & 31.33               \\ \hline

  \multicolumn{1}{|c|}{o4-mini}
    & \multicolumn{1}{c|}{Mimir, 10}
    & \multicolumn{1}{c|}{0.69}
    & \multicolumn{1}{c|}{0.84}
    & \multicolumn{1}{c|}{1.67}
    & 5.07                \\ \hline

  \multicolumn{1}{|c|}{GPT-4.1-mini}
    & \multicolumn{1}{c|}{Mimir, 10}
    & \multicolumn{1}{c|}{0.65}
    & \multicolumn{1}{c|}{0.81}
    & \multicolumn{1}{c|}{1.59}
    & 4.75                \\ \hline

  \multicolumn{1}{|c|}{GPT-4.1-mini}
    & \multicolumn{1}{c|}{Mimir, 30}
    & \multicolumn{1}{c|}{0.64}
    & \multicolumn{1}{c|}{0.81}
    & \multicolumn{1}{c|}{1.91}
    & 8.45                \\ \hline

  \multicolumn{1}{|c|}{GPT-4.1-mini}
    & \multicolumn{1}{c|}{10, No Val}
    & \multicolumn{1}{c|}{0.66}
    & \multicolumn{1}{c|}{0.85}
    & \multicolumn{1}{c|}{2.49}
    & 20.07               \\ \hline

  \multicolumn{1}{|c|}{GPT-4.1-mini}
    & \multicolumn{1}{c|}{10, No Meta}
    & \multicolumn{1}{c|}{0.53}
    & \multicolumn{1}{c|}{0.73}
    & \multicolumn{1}{c|}{2.09}
    & 4.57                \\ \hline

  \multicolumn{1}{|c|}{GPT-4.1-mini}
    & \multicolumn{1}{c|}{10, Naive RAG, Planner}
    & \multicolumn{1}{c|}{0.37}
    & \multicolumn{1}{c|}{0.79}
    & \multicolumn{1}{c|}{11.16}
    & 22.16               \\ \hline

  \multicolumn{1}{|c|}{GPT-4.1-mini}
    & \multicolumn{1}{c|}{10, Naive RAG}
    & \multicolumn{1}{c|}{0.17}
    & \multicolumn{1}{c|}{0.75}
    & \multicolumn{1}{c|}{5.53}
    & 9.99                \\ \hline
\end{tabular}
\end{adjustbox}
\end{table*}

\begin{figure*}[!hbt]
    \centering
    \includegraphics[width=0.8\textwidth]{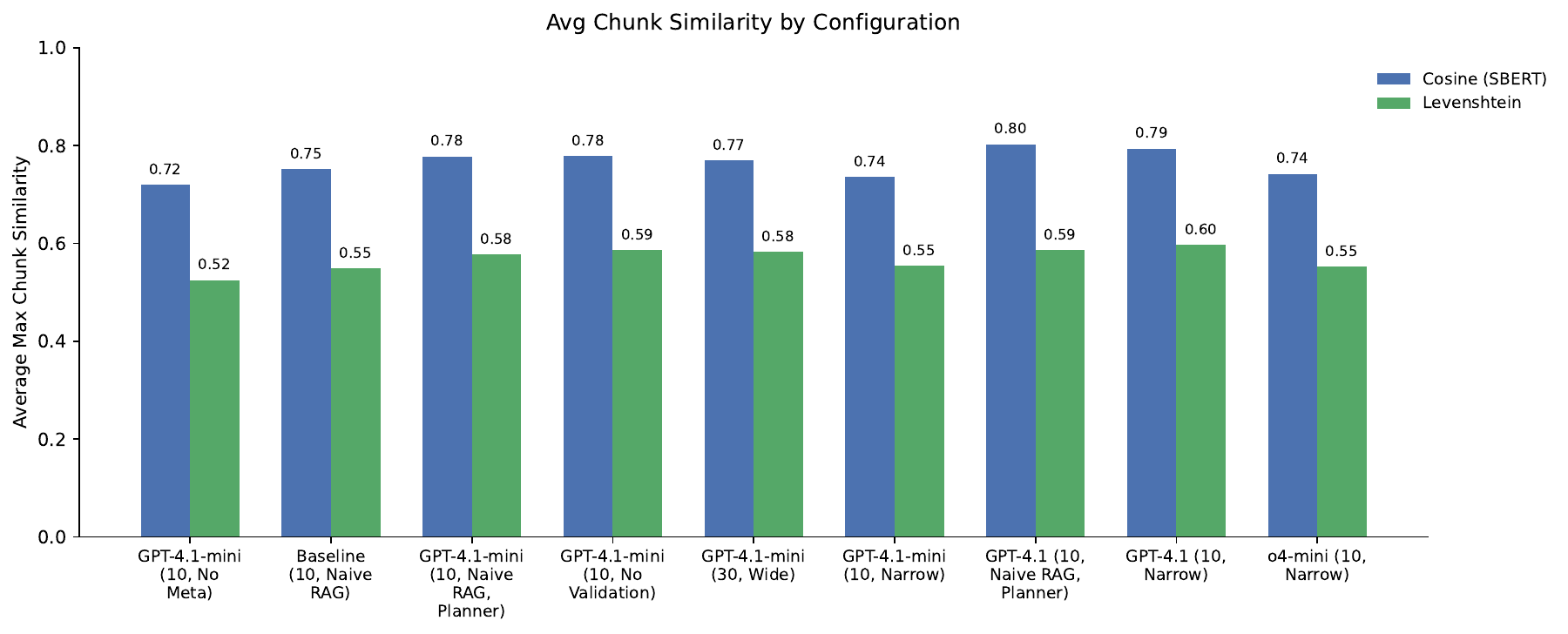} % Assuming path is correct
    \caption{Iteration 1: Average chunk similarity (Cosine and Levenshtein).}
    \label{fig:chunk_iter1_exp}
\end{figure*}

Figure~\ref{fig:chunk_iter1_exp} provides a deeper insight into the quality of the retrieved content by displaying the average maximum chunk similarity scores. The analysis reveals several key trends. The GPT-4.1 (10) configuration (full MimirRAG), which also achieved the highest end-to-end accuracy in Iteration 1 (82.7\%), correspondingly exhibited the highest chunk similarity scores on both metrics (Cosine 0.80, Levenshtein 0.59). This underscores the importance of retrieving semantically and textually aligned chunks for generating correct final answers. 

Furthermore, comparing MimirRAG configurations against Naive RAG variants using GPT-4.1-mini (Bar 2, 3, and 6 in Figure~\ref{fig:chunk_iter1_exp}) reveals an interesting pattern. The GPT-4.1-mini (10, Naive RAG, Planner) setup (Bar 3) achieved the highest average maximum Cosine similarity (0.78) among these GPT-4.1-mini based configurations, surpassing even the full MimirRAG GPT-4.1-mini (Mimir, 10) system (Bar 6: Cosine 0.74) on this specific retrieval metric. Despite this superior chunk similarity, the full MimirRAG system delivered significantly better end-to-end accuracy (76.0\% vs. 69.3\% for `Naive RAG + Planner' and 65.3\% for basic `Naive RAG'). We speculate that while the Planner in a simplified Naive RAG context might effectively identify semantically similar content, this content may not always be optimally relevant or from the precisely correct source. For example, a highly similar passage from a prior year's report rather than the target year.

\subsubsection{Error Analysis by Reasoning Type}
\begin{figure*}[!hbt]
 \centering
 \includegraphics[width=.8\textwidth]{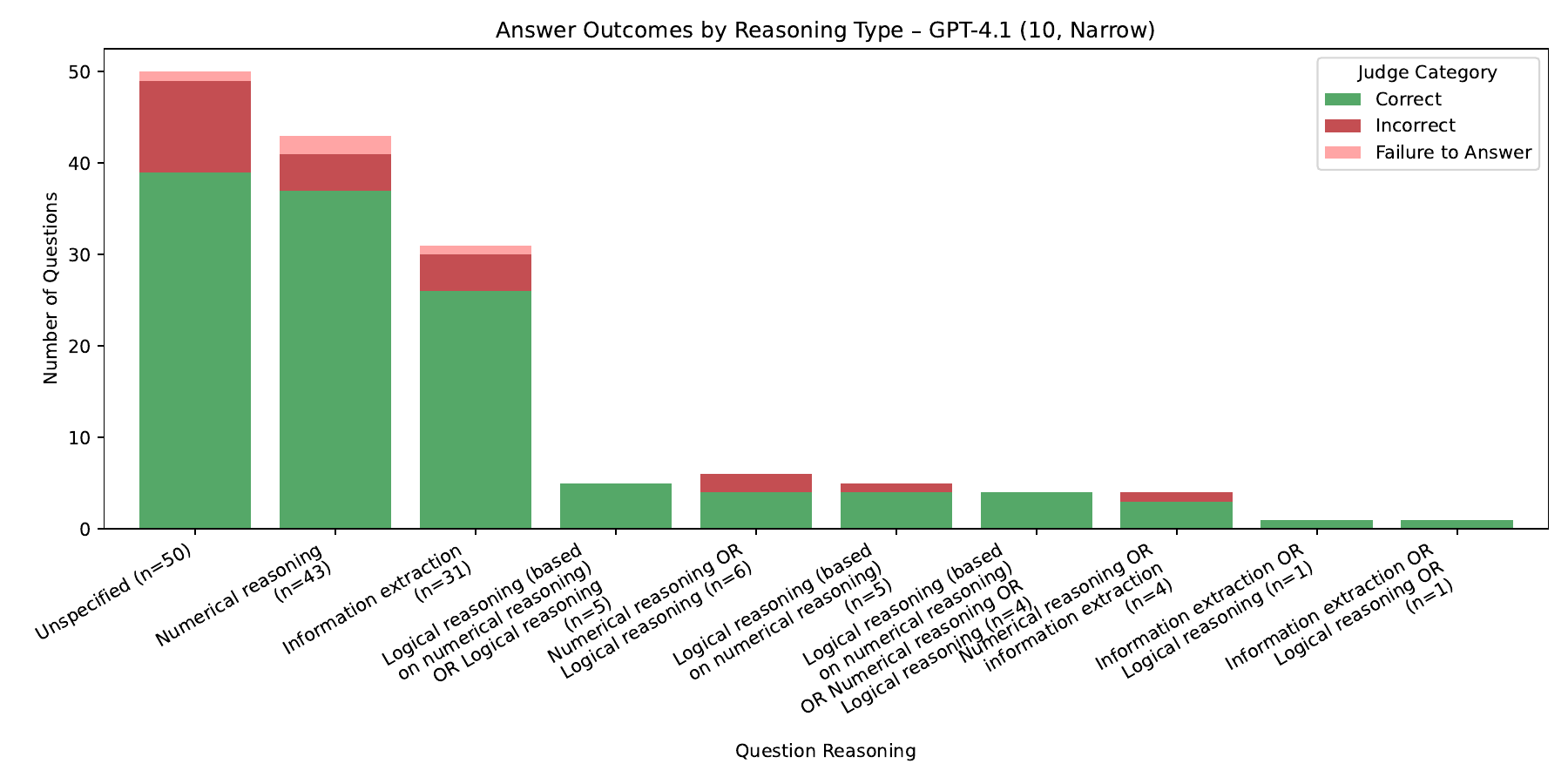}
 \caption{Iteration 1: Answer outcomes by reasoning type for GPT-4.1 (10).}
 \label{fig:reasoning-outcomes-exp}
\end{figure*}

An error analysis of the best-performing configuration (GPT-4.1, Mimir, 10) revealed that `Unspecified' (novel-generated) questions had the most incorrect answers (Figure~\ref{fig:reasoning-outcomes-exp}). We manually checked each error and found that many were due to numerical errors, but also related to partly missing data. This was often due to tables that had been split across chunks. If only the first part of the table was in the context, the Writer agent would try to answer the question based on that partial table, and often make estimates on the missing data. This means that some of the numerical errors are due to missing data, and not necessarily a numerical reasoning error, but rather an information retrieval problem. However, there were also errors due to direct numerical reasoning errors. A small number of errors were also due to poorly specified questions in the benchmark, for example:~\textit{financebench\_id: 00601}: ``What drove the reduction in SG\&A expense as a percent of net sales in FY2023?'' Missing the relevant company name, this is impossible with a shared vector store.~\textit{financebench\_id: 00822}: ``Were there any board member nominees who had substantially more votes against joining than the other nominees?'' Missing both the company name and the relevant year.

\subsubsection{Lessons Learned from first Iteration}
Iteration 1 revealed several key insights that directly informed refinements in Iteration 2: \textit{1)}~model size mattered, with GPT-4.1 generally outperforming smaller variants; \textit{2)} metadata improved performance through better precision in hierarchical search; \textit{3)} a low `Failed to Answer' rate suggested effective data retrieval in most cases;~\textit{4)} there were diminishing returns for chunk count initially, as scaling from 10 to 30 chunks for GPT-4.1-mini offered little accuracy gain without better ranking; \textit{5)}~numerical reasoning emerged as a key challenge; and~\textit{6)} table chunking issues, where split tables led to incomplete context for the Writer agent, were identified.
%

%================================================
\subsection{Second Iteration}
\label{sec:iter2}
%================================================
The benchmark methodology remained consistent with Iteration 1 (FinanceBench 150-question subset, Judge agent evaluation). This iteration aimed to improve two identified weaknesses.

\begin{enumerate}
    \item \textbf{Calculator Tool for Writer agent:} To address arithmetic errors observed in numerical reasoning tasks. This was implemented using \texttt{sympy}, allowing the agent to symbolically parse mathematical formulas, substitute variables with their corresponding values, evaluate the expressions numerically, and obtain results rounded to a specified precision.
    \item \textbf{Table-aware Chunking Strategy:} The Chonkie chunker was enhanced to better handle tables. After an initial recursive chunking based on Markdown structure (1800-character target, 300-character overlap), a specific table-merging step was introduced. This step identifies consecutive chunks where the first ends with a table row and the next begins with a table row (identified by lines starting with a pipe `\textbar'). Such chunks are merged together, up to a maximum length of 3600 characters (twice the standard chunk size), to prevent tables from being split across multiple, disconnected chunks. In effect, the strategy preserves a larger contiguous table fragment so that later retrieval returns complete financial rows more often. The original overlap refinement step was then applied to these potentially merged chunks.
\end{enumerate}

These targeted changes aimed to improve accuracy, particularly in numerical and table-based questions. To analyze the impact of the table-aware chunking strategy introduced, we examined the distribution of chunks per report. Without table merging (as in Iteration 1), the mean number of chunks per report was 431.6 (standard deviation: 284.6). With table merging enabled, the mean decreased to 380.2 chunks per report (standard deviation: 253.0). This reduction indicates that the table-merging step effectively consolidated content, leading to fewer, potentially more contextually complete, chunks for table-heavy documents.

%===============================================
\subsubsection{Performance of Models}
\label{sec:iter2-results}
%===============================================
Table~\ref{tab:model-performance-iter2} presents the performance of the improved MimirRAG. All configurations showed improvement. Notably, GPT-4.1 with 20 chunks achieved 89.3\% accuracy, surpassing its Iteration 1 best (82.7\%) and the original 85\% FinanceBench Oracle. This is approaching our updated GPT-4.1 Oracle baseline of 93.3\%, demonstrating the efficacy of the enhancements. The Naive RAG configuration (GPT-4.1-mini) also saw a performance boost of approximately 3 percentage points (from 65.3\% to 68.0\%) from the improved table-aware chunking (no calculator in the Naive approach).

GPT-4.1-mini also benefited significantly: its 10-chunk configuration accuracy rose from 76.0\% (Iteration 1, full MimirRAG) to 79.3\% (Iteration 2, full MimirRAG) with the new chunking and calculator tool. Increasing the chunk depth to 20 in this improved system further boosted its accuracy to 84.7\%, with a very low `Failed to Answer' rate (0.7\%). Unlike Iteration 1, scaling search breadth (chunk count) now yielded more substantial gains for GPT-4.1-mini.

\begin{table*}[!hbt]
\centering
\caption{Iteration 2: Performance comparison with system improvements (table-aware chunking and calculator tool).}
\label{tab:model-performance-iter2}
\begin{adjustbox}{width=.8\textwidth}
\begin{tabular}{|c|c|c|c|c|c|}
\hline
\textbf{Model} & \textbf{Configuration} & \textbf{Correct}      & \textbf{Incorrect}   & \textbf{Failed to answer} & \textbf{Rel. Cost (\$/\% Acc.)} \\ \hline
GPT-4.1-mini   & 10, Naive RAG          & 102 (68\%)            & 44 (29.3\%)          & 4 (2.6\%)                 & 0.029                           \\ \hline
GPT-4.1-mini   & Mimir, 10                     & 119 (79.3\%)          & 25 (16.7\%)          & 6 (4.0\%)                 & 0.025                           \\ \hline
GPT-4.1-mini   & Mimir, 20                     & 127 (84.7\%)          & 22 (14.7\%)          & \textbf{1 (0.7\%)}        & 0.024                           \\ \hline
GPT-4.1        & Mimir, 20                     & \textbf{134 (89.3\%)} & \textbf{15 (10.0\%)} & \textbf{1 (0.7\%)}        & 0.112                           \\ \hline
\end{tabular}
\end{adjustbox}
\end{table*}
\subsubsection{Retrieval Analysis}
\label{sec:iter2-retrieval-analysis}
At the document retrieval level, the core filtering and document search mechanisms remained consistent with Iteration 1, the results are shown in Table~\ref{tab:doc-performance-iter2}. At the document level, Hit@1 performance showed varied changes. For the GPT-4.1 model, increasing the retrieved chunks from 10 (Iteration 1, Hit@1: 0.67) to 20 (Iteration 2, Hit@1: 0.65) saw a slight decrease in Hit@1. For GPT-4.1-mini, comparing the 10-chunk configuration between Iteration 1 (Hit@1: 0.65) and Iteration 2 (Hit@1: 0.65) showed no change in Hit@1. However, within Iteration 2, increasing the chunk depth for GPT-4.1-mini from 10 to 20 significantly improved Hit@1 from 0.65 to 0.73. This latter configuration also surpassed the Hit@1 achieved by GPT-4.1-mini with 30 chunks in Iteration 1 (0.64). The Hit@5 metric remained relatively stable or showed slight improvements for comparable MimirRAG configurations.

The average number of distinct documents processed per question (`Avg docs') varied across configurations, with some Iteration 2 setups (e.g., GPT-4.1-mini with 10 or 20 chunks) processing slightly more unique documents on average than their direct Iteration 1 counterparts. The average number of chunks sent to the Writer Agent (`Avg chunks') is now influenced by both the number of initially retrieved chunks and the effects of the table-aware chunking strategy, as discussed; some chunks are now larger and contain more information.

\begin{table*}[!hbt]
\centering
\caption{Iteration 2: Retrieval performance across configurations.}
\label{tab:doc-performance-iter2}
\begin{adjustbox}{width=.8\textwidth}
\begin{tabular}{|c|c|c|c|c|c|}
\hline
\multicolumn{1}{|c|}{\textbf{Model}}
  & \multicolumn{1}{c|}{\textbf{Setting}}
  & \multicolumn{1}{c|}{\textbf{Hit@1}}
  & \multicolumn{1}{c|}{\textbf{Hit@5}}
  & \multicolumn{1}{c|}{\textbf{Avg docs}}
  & \textbf{Avg chunks} \\ \hline
\multicolumn{1}{|c|}{GPT-4.1}
  & \multicolumn{1}{c|}{Mimir, 20}
  & \multicolumn{1}{c|}{0.65}
  & \multicolumn{1}{c|}{0.85}
  & \multicolumn{1}{c|}{2.01}
  & 9.25                \\ \hline
\multicolumn{1}{|c|}{GPT-4.1-mini}
  & \multicolumn{1}{c|}{Mimir, 20}
  & \multicolumn{1}{c|}{0.73}
  & \multicolumn{1}{c|}{0.86}
  & \multicolumn{1}{c|}{1.89}
  & 7.53                \\ \hline
\multicolumn{1}{|c|}{GPT-4.1-mini}
  & \multicolumn{1}{c|}{Mimir, 10}
  & \multicolumn{1}{c|}{0.65}
  & \multicolumn{1}{c|}{0.85}
  & \multicolumn{1}{c|}{1.79}
  & 5.29                \\ \hline
\multicolumn{1}{|c|}{GPT-4.1-mini}
  & \multicolumn{1}{c|}{Naive RAG}
  & \multicolumn{1}{c|}{0.25}
  & \multicolumn{1}{c|}{0.79}
  & \multicolumn{1}{c|}{5.58}
  & 10.00               \\ \hline
\end{tabular}
\end{adjustbox}
\end{table*}

\begin{figure*}[!hbt]
    \centering
    \includegraphics[width=.8\textwidth]{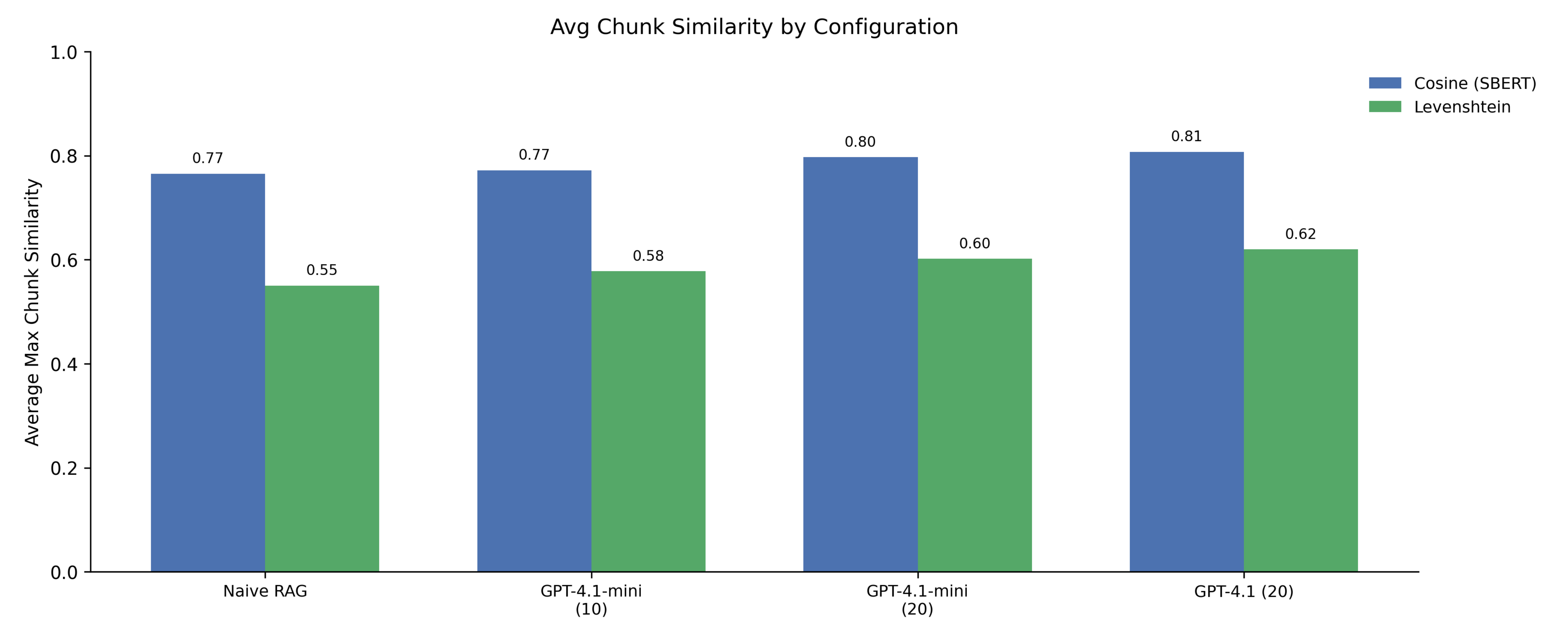}
    \caption{Iteration 2: Average chunk similarity (Cosine and Levenshtein).}
    \label{fig:chunk-similarity-iter2}
\end{figure*}

Figure~\ref{fig:chunk-similarity-iter2} shows the chunk similarity scores for Iteration 2. The best performing model, GPT-4.1 (Mimir, 20), achieved the highest average Cosine and Levenshtein similarity scores, suggesting that the retrieved chunks were semantically closer and more textually similar to the gold-standard evidence.

\subsubsection{Error Analysis by Reasoning Type}
\label{sec:iter2-error-analysis}
An error analysis for the best-performing configuration in Iteration 2 (GPT-4.1, Mimir, 20) is presented in Table~\ref{tab:question-type-errors-iter2}.
The introduction of the calculator tool and table-aware chunking aimed to reduce errors, particularly in numerical and table-based questions.
\begin{table*}[!hbt]
\centering
\caption{Iteration 2: Error breakdown by question type for GPT-4.1 (20).}
\label{tab:question-type-errors-iter2}
\begin{tabular}{|c|c|c|c|}
\hline
\textbf{Question Type} & \textbf{Correct} & \textbf{Failure to Answer} & \textbf{Incorrect} \\ \hline
Domain-relevant        & 45               & 1                          & 4                  \\ \hline
Metrics-generated      & 47               & 0                          & 3                  \\ \hline
Novel-generated        & 42               & 0                          & 8                  \\ \hline
\end{tabular}
\end{table*}
Interestingly, while overall accuracy improved significantly in Iteration 2, an analysis revealed that six questions which were answered correctly by the top configuration in Iteration 1 were \textit{answered incorrectly in Iteration 2 by its respective top configuration}. This suggests that while the system improvements were broadly beneficial, they may have introduced regressions for a small subset of cases. For two of these, we see that this is due to different judgment from the Judge agent and not a difference in response from MimirRAG. This highlights a potential source of variance in automated evaluation. The remaining four regressions appear to be genuine changes in system output. Further investigation into these specific instances is warranted and will be discussed in Section~\ref{sec:discussion:interpretation}. If these six questions had been answered correctly, the accuracy for GPT-4.1 (Mimir, 20) would have reached 93.3\%.

\subsubsection{Lessons Learned from second Iteration}
Finally, we learned from second iteration: \textit{1)}~Targeted improvements yield significant gains: The calculator tool and table-aware chunking directly addressed weaknesses from Iteration 1, leading to substantial accuracy increases across models.~\textit{2)}~Surpassing original Oracle baseline: MimirRAG (GPT-4.1, Mimir, 20), with 89.3\% accuracy, surpassed the original 85\% FinanceBench Oracle (GPT-4-Turbo) and closely approached our more up-to-date Oracle baseline (93.3\% with GPT-4.1).~\textit{3)}~Increased chunk depth becomes beneficial: with better underlying processing (especially more complete table chunks), providing more context became more effective, unlike the diminishing returns seen in Iteration 1 for smaller models.~\textit{4)}~Reduced errors and failures: the `Incorrect' and `Failed to Answer' rates dropped, indicating more robust retrieval and downstream reasoning.

%%%%%%%%%%%%%%%%%%%%%%%%%%%%%%%%%%%
\section{Expert Validation}
\label{sec:human-eval-overview}
%%%%%%%%%%%%%%%%%%%%%%%%%%%%%%%%%%%
This section details an iterative evaluation process, beginning with an initial prototype assessment by one expert (P1), leading to system refinements, followed by a broader evaluation with four financial analysts (P1-P4). The data collection process followed a mixed-method strategy i.e., qualitative data from interviews and open-ended survey responses were thematically coded (e.g., for workflow impact, trust calibration, feature requests), while quantitative data from Likert-scale surveys (A/B test in Iteration 1, feedback survey in Iteration 2) provided descriptive statistics on perceived system attributes.

\subsection{Iteration 1: Formative Evaluation with Participant 1}
The initial prototype was evaluated by one financial analyst (P1) through an A/B test against Microsoft Copilot and a semi-structured interview. Microsoft Copilot was selected as it is the current alternative at the bank and is based on GPT-4o.
Table~\ref{tab:ab_means} presents the mean scores from the A/B test, where P1 evaluated MimirRAG and Copilot on five financial questions. MimirRAG outperformed Copilot across all five dimensions, notably in relevance (+1.0) and usefulness (+0.6) on a five-point Likert scale.
\begin{table}[!hbt]
\centering
\caption{Average analyst ratings (1 = poor, 5 = excellent) (N=1)}
\label{tab:ab_means}
\begin{tabular}{|c|c|c|}
\hline
\textbf{Dimension}    & \textbf{Copilot} & \textbf{MimirRAG} \\ \hline
Relevance to query    & 3.2              & \textbf{4.2}      \\ \hline
Factual correctness   & 4.0              & \textbf{4.2}      \\ \hline
Usefulness            & 3.0              & \textbf{3.6}      \\ \hline
Clarity and reasoning & 3.6              & \textbf{4.0}      \\ \hline
Source traceability   & 4.4              & \textbf{4.6}      \\ \hline
\end{tabular}
\end{table}
Qualitative feedback from P1, gathered through the interview and A/B test comments, provided richer context. P1 found MimirRAG's answers more comprehensive and historically aware than Copilot's, and valued the chat interface for iterative questioning. While initial impressions of MimirRAG were positive, P1 emphasized that sustained trust would require consistent and accurate performance over time, given the potential impact of errors in their professional recommendations, and that he would always review outputs manually rather than expecting a `plug-and-play' solution.

The most actionable insight from this initial evaluation was P1's strong emphasis on integrating personal research notes to create a `private knowledge vault', identified as a key unique feature. This feedback, along with the generally positive reception and the noted importance of trust and consistency, guided key enhancements implemented before Iteration 2. First, a \textit{Notes Integration Feature} was developed, providing a system for uploading, managing, and querying personal notes, which included processing P1's own notes for testing. Second, to gather broader perspectives, the \textit{number of evaluators was expanded to include three additional analysts} for the subsequent iteration.

\subsection{Analysts Feedback Survey}
Following system enhancements, the prototype was evaluated by four financial analysts (P1-P4). Feedback was gathered through a survey incorporating Likert-scale questions and open-ended responses, supplemented by semi-structured follow-up interviews with P2, P3, and P4. P1 also provided specific feedback on the newly integrated notes feature.

\subsubsection{Quantitative Findings}
The Likert-scale responses (where 1 = Worst/Least, 5 = Best/Most) provided the following insights (Table~\ref{tab:iter2_survey_ratings_detailed}):
\begin{table*}[!hbt]
\small
\centering
\caption{Analyst Survey Ratings (N=4)}
\label{tab:iter2_survey_ratings_detailed}
\begin{adjustbox}{width=.8\textwidth}
\begin{tabular}{|c|c|c|c|c|c|}
\hline
\textbf{Question}                   & \textbf{P1} & \textbf{P2} & \textbf{P3} & \textbf{P4} & \textbf{Mean} \\ \hline
3. How helpful is the system in retrieving info?  & 3           & 4           & 3           & 5           & \textbf{3.75} \\ \hline
4. How clear/transparent is the reasoning? & 4           & 4           & 4           & 5           & \textbf{4.25} \\ \hline
5. To what extent do you trust the system?     & 4           & 3           & 4           & 4           & \textbf{3.75} \\ \hline
6. Would the system potentially help you to save time?          & 4           & 4           & 3           & 5           & \textbf{4.00} \\ \hline
\end{tabular}
\end{adjustbox}
\end{table*}

The system's transparency was rated highly (Mean: 4.25), with good potential for time-saving (Mean: 4.00). Helpfulness and trust scored positively (Mean: 3.75 each). As elaborated in the qualitative findings, trust was noted as needing to be earned over consistent use.

\subsubsection{Qualitative Findings}
Qualitative insights (gathered from survey and interviews), synthesized from open-ended survey responses and follow-up interviews (P2-P4), and P1's feedback on the notes feature, revealed several key themes. Users valued features like the Notes search (P1), ease of navigation (P2), qualitative data retrieval (P3), and system speed (P4). Key improvement areas identified included expanding data access (P1, P2), improving document-level control (P2), better handling of tables/graphs (P3), and enhanced output formats such as charts or spreadsheet integration (P4).

Willingness to integrate the tool into regular workflows varied: P4 was open, P1 and P3 were tentative, and P2 was hesitant, citing the richer feature sets in commercial tools. Regarding accuracy, P2 and P4 reported strong performance for factual queries and direct document retrieval with minimal hallucination. In contrast, P3 encountered difficulties with table and figure extraction, though text summarization worked well. Some participants mentioned reverting to other tools for certain tasks or when encountering perceived inconsistencies. Trust was consistently described as evolving and highly sensitive to performance; P4 considered the system trustworthy but fragile, while P3's mixed experience led to caution, emphasizing that consistent functionality across all data types is essential, often leading to cautious initial use and manual verification of outputs. A recurring theme was the critical need for broader data integration (e.g., live news, financial, transcripts) and improved processing capabilities to meet professional expectations.

\subsubsection{Way Forward}
The Iteration 2 evaluation with four analysts confirmed the value of the personal notes integration (particularly enhancing usability for P1, who tested this feature). Overall feedback from surveys and interviews was positive regarding helpfulness, transparency, and time-saving potential. However, two key areas for future improvement emerged: \textit{1)} Trust, which hinges on consistent performance, especially with structured data. \textit{2)} The need for expanded data integration and processing capabilities to align with professional tools and expectations.

%%%%%%%%%%%%%%%%%%%%%%%%%%%%%%%%%%%%
\section{Hindsight}
\label{chap:discussion}
%%%%%%%%%%%%%%%%%%%%%%%%%%%%%%%%%%%%
%----------------------------------------------------------
\subsection{Interpretation of the Findings}
\label{sec:discussion:interpretation}

\subsubsection{Findings from Iteration 1}
Ablation studies in Iteration 1 revealed that individual components of the MimirRAG pipeline significantly affect accuracy. Even the reimplemented Naive RAG using GPT-4.1-mini, combined with Docling's PDF-to-Markdown conversion and structure-aware chunking, achieved a solid baseline accuracy of 65.3\%. Introducing a Planner-based query expansion improved performance to 69.3\%, with a notable increase in document retrieval accuracy (Hit@1 from 0.17 to 0.37), driven by better initial queries.

In contrast, the No Meta configuration—representing hierarchical search without metadata filtering—underperformed (62.0\% accuracy), likely due to semantic ambiguity in document summaries. However, integrating hierarchical search with metadata filtering significantly boosted performance, raising accuracy to 76.0\% and Hit@1 to 0.65. It also reduced the average number of retrieved documents (from 14.23 to 2.02), minimizing noise and improving precision. This highlights the critical role of metadata in enhancing retrieval efficiency and addresses a key gap in existing financial RAG systems.

Iteration 1 showed that the Validator agent significantly reduced noise at the chunk level, improving accuracy by 4.7\% (from 71.3\% without Validator to 76.0\% with it). Although the No Validator setup yielded slightly higher chunk similarity scores, it resulted in less accurate answers—indicating that the Validator's strength lies in filtering out irrelevant or contradictory content, thus enhancing final answer quality beyond what similarity metrics alone can capture. Additionally, targeted model allocation proved beneficial: using the stronger o4-mini model for the Planner and Writer (while keeping GPT-4.1-mini for Search and Validation) yielded a 3\% accuracy gain. Overall, results confirmed that system performance scales with LLM quality and highlighted the value of modular RAG design, allowing easy integration of more capable models in specific roles.

\subsubsection{Findings from Iteration 2}
During Iteration 2, MimirRAG continued to outperform baseline methods; for example, using GPT-4.1-mini, it achieved 84.7\% accuracy (20 chunks) compared to Naive RAG's 68.0\% (Table~\ref{tab:model-performance-iter2}). The advantage of stronger models remained clear, with GPT-4.1 reaching 89.3\% accuracy, surpassing GPT-4.1-mini (84.7\%). Iteration 2 addressed specific weaknesses from Iteration 1, including numerical reasoning and table fragmentation. The introduction of a calculator tool for the Writer Agent and a table-aware chunking strategy (`Chonkie Chunker') led to measurable improvements. Table-aware chunking alone improved Naive RAG performance by approximately 2.7\%, increasing GPT-4.1-mini (10, Naive RAG) accuracy from 65.3\% to 68.0\%, even without the calculator tool.

A key insight was that expanding retrieval breadth became more beneficial in Iteration 2. While increasing from 10 to 30 chunks showed no gain in Iteration 1 (both at 76.0\%, Table~\ref{tab:model-performance-iter1}), raising chunks from 10 to 20 in Iteration 2 led to a significant jump for GPT-4.1-mini—from  79.3\% to 84.7\% (Table~\ref{tab:model-performance-iter2}). This suggests that improved chunking—especially merged table segments—produced more useful retrievals because the Writer agent more often received complete numerical context rather than fragmented tables. However, these gains introduced new complexities. Six questions previously answered correctly in Iteration 1 were marked incorrect in Iteration 2’s top MimirRAG configuration. In two cases, answers were nearly identical but received different judgments from the LLM-based Judge agent; the remaining four were true regressions. This reflects known issues in FinanceBench, where other researchers have also noted evaluation inconsistencies~\cite{nguyen2024enhancing,vectify2025mafingithub}. These findings highlight the trade-offs in RAG system optimization and the limitations of relying solely on automated evaluation.

\subsubsection{Findings from Expert Feedback}
\label{sec:discussion:expert_validation}
Our iterative expert validation with financial analysts underscored that successful RAG tool integration hinges on a triad of interconnected factors: earned trust, comprehensive data integration, and deep personalization. The critical importance of \textit{trust calibration}~\cite{lee2004trust} was evident. While MimirRAG's low hallucination rates were valued and initial quantitative trust was moderate (Mean: 3.75/5), qualitative feedback across iterations revealed trust as an evolving state, earned through consistent and reliable performance. Analysts, aware of the high-stakes nature of their work, approached the system cautiously, often verifying results manually (a behavior P1 highlighted and others echoed). This cautious adoption and dynamic trust adjustment based on task-specific performance (e.g., P3's issues with structured data) sometimes led to reverting to familiar tools, aligning with established trust models in automation~\cite{lee2004trust,shneiderman2020human} where users avoid over-reliance until dependability is proven.

Next, the \textit{scope and integration of data} significantly impacted perceived utility. While MimirRAG was useful with uploaded documents, analysts (notably P1, P2, P4) consistently stressed the need for broader, dynamic data sources and advanced numerical processing. The current static corpus was seen as limiting efficiency for real-time tasks, prompting the use of alternative platforms. This highlights a key challenge: future specialized RAG systems require architectural flexibility for robust, low-latency integration of heterogeneous live data, moving beyond curated knowledge bases to genuinely support dynamic professional data ecosystems.

Finally, the enthusiastic reception of the `Notes Integration' feature, driven by P1's desire for a `private knowledge vault', demonstrated the critical role of \textit{personalization}. For expert users, seamlessly querying and integrating their curated insights with system-retrieved information is a core value proposition, not just a convenience. This suggests that future RAG systems should prioritize architectures enabling secure and efficient personal knowledge base integration, offering a key differentiator by transforming them into customized analytical partners. In essence, developing effective RAG tools for professionals demands more than advanced retrieval. It requires a user-centric approach that holistically addresses the earned nature of trust, the necessity of comprehensive data integration, and the significant value of deep personalization to achieve genuine workflow integration.

\subsection{Implications of the Results}
The findings of this research carry important implications for various stakeholders and the broader understanding of RAG in finance.

\subsubsection{Practical Implications}
\textit{1)}~For Developers \& System Designers:~Our findings validate specific design principles derived from MimirRAG: domain-appropriate, table-aware chunking, robust metadata filtering for targeted retrieval, layered agentic processes with validation, and user-centric design for trust and personalization. These are critical for building effective financial RAG tools. \textit{2)}~For Financial Analysts:~MimirRAG demonstrates a path to reduced information overload, enabling a shift from data gathering to higher-value interpretation, as suggested by their feedback on potential time savings and the utility of features like `Notes Integration'. \textit{3)}~Empirical Validation of Metadata's Crucial Role:~This study provides strong empirical validation for metadata's crucial role in enhancing RAG precision (e.g., Iteration 1 Hit@1 improvement from 0.53 to 0.65 with metadata), a principle highly generalizable to other domains with semi-structured corpora (e.g., legal, medical).

\subsubsection{Theoretical Implications}
This research advances the field of \textit{1)}~Human-AI Collaboration by providing empirical insights into trust calibration in high-stakes professional settings. Analysts exhibited `rational under-utilization', gradually building trust through consistent performance, a pattern likely applicable in other critical domains.~\textit{2)}~The study also empirically validates the Modular RAG paradigm in finance, demonstrating its advantages over simpler RAG setups through tailored, task-specific processing. Iterative improvements in the MimirRAG prototype highlight the power and adaptability of this modular approach. Although designed for financial analysts at a Danish bank, MimirRAG's agentic, modular architecture, featuring enhanced query handling, targeted error mitigation, and tool integration suggests broad applicability. This design is especially suited for other specialized domains that require complex, multi-step retrieval and synthesis workflows, where generic RAG systems may be insufficient. Adapting this approach would require domain-specific tuning of agents and tools, but the core design philosophy most likely remains transferable.

\subsubsection{Societal Implications and Considerations}
\textit{1)}~Socio-Technical Shifts:~Beyond the direct benefits to financial analysts, the deployment of advanced RAG systems like MimirRAG also asks for reflection on broader societal shifts. The increasing automation of information-intensive tasks may reshape the skillsets required in the financial labor market, potentially shifting human analysts towards more strategic and interpretive roles while requiring ongoing adaptation and training. \textit{2)}~Ethical Governance and Responsible AI: as AI tools become more essential in financial analysis, ensuring proper ethical governance and addressing potential biases inherent in models or data becomes important to maintain market integrity and public trust. While MimirRAG is designed to augment human expertise, the wider adoption of such technologies calls for a proactive approach to these societal and ethical dimensions, particularly with emerging regulations like the EU AI Act~\cite{europarl2023AIAct}.

\subsection{Limitations}
\label{sec:discussion:limitations}
While this research offers important insights into RAG-based solutions for financial analysts, its findings should be interpreted with consideration of limitations, specifically related to the benchmark scope, the nature of expert validation, and constraints of the prototype system.

At the document-processing level, an additional limitation is the imperfect recovery of structure from PDFs. Although our parsing stack preserves layout and tables where possible, OCR and table extraction can still introduce errors in scanned or visually complex filings, which in turn can affect chunking and downstream retrieval.
\begin{enumerate}
    \item \textit{Benchmark Experiments:}~Several limitations should be considered when interpreting these findings. First, the evaluation is based on the 150-question open-source subset of the FinanceBench dataset. While useful for detailed analysis, results may not generalize to the full, non-public dataset, which we were unable to access. Additionally, our study relies on a single benchmark. Although other benchmarks like FinDER~\cite{choi2025finder} emerged during the research, their datasets were not publicly available for comparison. We also identified data quality issues within the FinanceBench subset, such as missing information (e.g., company names, fiscal years), incorrect gold answer calculations, and ambiguity in questions that allow multiple valid answers depending on financial ratio conventions. These concerns have also been reported by others~\cite{vectify2025mafingithub}.
    \item  \textit{OpenAI GPT-focused:}~Our experiments were limited to LLMs from OpenAI. A broader investigation, including other proprietary models and a diverse range of open-source alternatives, could provide a more comprehensive understanding of how different model architectures and training methodologies impact RAG performance in the financial domain.
    \item  \textit{Judge Agent:}~The evaluation of system responses was conducted using an LLM-based Judge agent. Although this is an increasingly common practice in NLP research~\cite{Zheng2023JudgingLWA,gu2024survey} and we try to mitigate known self-bias problem~\cite{Xu2024PrideAPA,ye2024justiceprejudicequantifyingbiases} by using a different model (GPT-4o) for judging than those used in the RAG pipeline, LLM-based evaluation can have inherent limitations and may not perfectly replicate human judgment.
    \item  \textit{Prompt engineering:}~Our prompt engineering was largely ad hoc. While we iteratively refined prompts based on observed performance, we did not employ systematic prompt optimization frameworks such as DSPy~\cite{khattab2023dspy}, which could potentially unlock further performance gains through more structured prompt engineering.
\end{enumerate}
In addition, the expert validation had several limitations. It involved a small (four) sample of financial analysts, all from a single case company, which limits the generalizability of findings on usability, trust, and workflow integration. Participant interactions with the MimirRAG prototype were relatively brief (e.g., two weeks for P1), potentially missing longer-term dynamics. The user interface prioritized functional evaluation over user experience design, which may have influenced perceptions, while systematically comparing different UI/UX (user experience) for financial RAG tools would yield valuable design guidelines. Additionally, specific features like `search-my-notes' were only tested by one participant (P1), limiting broader conclusions about their effectiveness. Research should explore how the increasing abilities of AI tools (such as RAG systems) might reshape the core skill sets required of financial analysts, while further investigation is needed into which explainability and interpretability techniques are most effective and valued by financial analysts.

\subsection{Lessons Learned}
Beyond the implemented enhancements, Iteration 2 also explored additional improvements that were ultimately excluded from the final system due to limited performance benefits or added complexity on the benchmark dataset. 
\begin{enumerate}
    \item First, inspired by~\cite{Anthropic_2025}, a `think' tool was tested for the Planner and Writer agents to support internal monologue and structured reasoning before complex tool use or generation. However, in our configuration, this addition did not improve performance. To preserve system simplicity, it was not included in the final Iteration 2 setup.
    \item  Second, we experimented with giving the Search agent finer control over retrieval by splitting the hybrid search into two tools: search reports and search chunks. The agent would first filter documents using search reports and then perform chunk-level retrieval using selected document IDs. Although this approach improved the document-level Hit@1 metric, it introduced latency and did not result in higher overall accuracy on FinanceBench. Therefore, the simpler unified search approach was retained.
\end{enumerate}
While not included in the final system due to limited benchmark gains, the split search strategy may offer practical advantages in real-world scenarios. Unlike benchmarks, which ensure document availability, live systems may encounter missing documents (e.g., a SEC F-20 form for year 2023 filing). In such cases, a more autonomous Search agent, capable of adaptive strategies like expanding date ranges or checking alternate years, could improve retrieval. Although this behavior wasn't formally evaluated, it underscores the potential value of agent autonomy in production environments.

\subsection{Ethical Aspect}
Deploying powerful AI like RAG in high-stakes financial decision-making demands considering key ethical dimensions. Research is critically needed to investigate and mitigate potential biases embedded in the training data for both LLMs and financial embedding models, as these can affect RAG outputs, and to establish clear lines of accountability for any erroneous or biased results. Furthermore, developing robust frameworks for responsible AI governance and regulation within the financial industry is crucial. Such frameworks must address transparency, security, and data privacy and ensure compliance with emerging legislation, such as the European AI Act~\cite{europarl2023AIAct}.
%

%%%%%%%%%%%%%%%%%%%%%%%%%%%%%%%%%%%
\section{Conclusion}
\label{chap:conclusion}
%%%%%%%%%%%%%%%%%%%%%%%%%%%%%%%%%%%
This work investigated the effective design and integration of RAG-based solutions for extracting meaningful financial insights, with a specific focus on enhancing the information retrieval workflows of financial analysts. Through the iterative design, development, and mixed-methods evaluation of our prototype system, MimirRAG, this research offers both practical architectural considerations and crucial insights into the human-centric factors affecting the adoption of such AI tools in the high-stakes financial domain. Our findings directly address our subresearch questions. MimirRAG's modular, multi-agent architecture (incorporating robust pre-retrieval, advanced agentic retrieval, and domain-specific generation) demonstrates high effectiveness for financial information search tasks (first subresearch question). From its evaluation, we derived key design principles (second subresearch question), including the critical use of domain-appropriate chunking, the significant impact of metadata integration, the benefits of layered retrieval with validation, and the necessity of designing for trust and seamless workflow integration. Furthermore, our expert validation demonstrates that successful adoption hinges on establishing \textit{calibrated trust}, ensuring \textit{comprehensive data integration}, and providing \textit{personalized augmentation of expertise} (third subresearch question). These findings collectively suggest that meaningful financial insights can be effectively achieved when cutting-edge technology is combined with human-centric design. Such extraction extends beyond improved data retrieval, requiring a system that processes complex financial data accurately, presents findings transparently, and supports iterative exploration. Through its demonstrated capabilities and positive expert feedback, MimirRAG serves as a compelling proof-of-concept that RAG technology holds significant promise for transforming information workflows in financial analysis. 

To fully realize this potential, future development should prioritize the fine-tuning and evaluation of embedding models and re-rankers specifically on financial corpora, an important next step for improving domain-specific performance. Two key areas, identified through analyst feedback, stand out for enhancing tool value: first, integrating live data sources (news, full financial reports, transcripts) to expand the system's knowledge base, and second, enabling more natural, context-aware generation. The field would also benefit from new, comprehensive benchmarks reflecting realistic analyst workflows, supporting broader reasoning types and multi-document scenarios to better evaluate advanced RAG capabilities. In parallel with these technical advancements, ongoing research must continue to explore the human factors critical for successful adoption and integration of RAG systems into professional practice.

\section*{Acknowledgment}
We sincerely thank the analysts who generously shared their time and expertise. Their candid feedback played a crucial role in shaping our experiments and reinforcing the validity of our conclusions.

%%%%%%%%%%%%%%%%%%%%%%%%%%%%%%%%%%
\section*{Declarations}
\subsection*{Use of Generative AI}
%%%%%%%%%%%%%%%%%%%%%%%%%%%%%%%%%%
Generative AI was used solely for proofreading (spelling, grammar, punctuation, and formatting issues) in this manuscript. It was not used for any conceptual, technical, or scholarly content, including figures, references, or code. All substantive work is the authors' own.

%%%%%%%%%%%%%%%%%%%%%%%%%%%%%%%%%%
\subsection*{Ethical Approval}
%%%%%%%%%%%%%%%%%%%%%%%%%%%%%%%%%%
Not applicable.

%%%%%%%%%%%%%%%%%%%%%%%%%%%%%%%%%%
\subsection*{Competing Interests}
%%%%%%%%%%%%%%%%%%%%%%%%%%%%%%%%%%
The authors declare that they have no known competing financial interests or personal relationships that could have appeared to influence the work reported in this paper.

%%%%%%%%%%%%%%%%%%%%%%%%%%%%%%%%%%
\subsection*{Authors' Contributions}
%%%%%%%%%%%%%%%%%%%%%%%%%%%%%%%%%%

\textbf{Magnus} Methodology, Formal Analysis, Investigation, Software, Visualization, Writing - Original Draft.
\textbf{Wilmer} Methodology, Formal Analysis, Investigation, Writing - Original Draft.
\textbf{Somnath} Conceptualization, Supervision, Visualization, Writing - Review \& Editing.
\textbf{Mansoor} Conceptualization, Resources, Supervision - Review \& Editing.
\textbf{Mikkel} Conceptualization, Writing - Review \& Editing.

%%%%%%%%%%%%%%%%%%%%%%%%%%%%%%%%%%
\subsection*{Funding}
%%%%%%%%%%%%%%%%%%%%%%%%%%%%%%%%%%
Not applicable.
%%%%%%%%%%%%%%%%%%%%%%%%%%%%%%%%%%
\subsection*{Availability of data and materials}
%%%%%%%%%%%%%%%%%%%%%%%%%%%%%%%%%%
Not applicable.
\bibliographystyle{elsarticle-num}
\bibliography{ref}

@article{singh2025agentic,
  title={Agentic Retrieval-Augmented Generation: A Survey on Agentic RAG},
  author={Singh, Aditi and Ehtesham, Abul and Kumar, Saket and Khoei, Tala Talaei},
  journal={arXiv preprint arXiv:2501.09136},
  year={2025}
}

@article{nie2024survey,
  title={A Survey of Large Language Models for Financial Applications: Progress, Prospects and Challenges},
  author={Nie, Yuqi and Kong, Yaxuan and Dong, Xiaowen and Mulvey, John M and Poor, H Vincent and Wen, Qingsong and Zohren, Stefan},
  journal={arXiv preprint arXiv:2406.11903},
  year={2024}
}

@article{impink2022regulation,
  title={Regulation-induced disclosures: evidence of information overload?},
  author={Impink, Joost and Paananen, Mari and Renders, Annelies},
  journal={Abacus},
  volume={58},
  number={3},
  pages={432--478},
  year={2022},
  publisher={Wiley Online Library}
}

@article{setty2024improving,
  title={Improving retrieval for rag based question answering models on financial documents},
  author={Setty, Spurthi and Thakkar, Harsh and Lee, Alyssa and Chung, Eden and Vidra, Natan},
  journal={arXiv preprint arXiv:2404.07221},
  year={2024}
}

@article{li2024alphafin,
  title={Alphafin: Benchmarking financial analysis with retrieval-augmented stock-chain framework},
  author={Li, Xiang and Li, Zhenyu and Shi, Chen and Xu, Yong and Du, Qing and Tan, Mingkui and Huang, Jun and Lin, Wei},
  journal={arXiv preprint arXiv:2403.12582},
  year={2024}
}

@misc{wu2023bloombergGPT,
      title={BloombergGPT: A Large Language Model for Finance}, 
      author={Shijie Wu and Ozan Irsoy and Steven Lu and Vadim Dabravolski and Mark Dredze and Sebastian Gehrmann and Prabhanjan Kambadur and David Rosenberg and Gideon Mann},
      year={2023},
      eprint={2303.17564},
      archivePrefix={arXiv},
      primaryClass={cs.LG},
      url={https://arxiv.org/abs/2303.17564}, 
}

@article{liu2023fingpt,
  title={Fingpt: Democratizing internet-scale data for financial large language models},
  author={Liu, Xiao-Yang and Wang, Guoxuan and Yang, Hongyang and Zha, Daochen},
  journal={arXiv preprint arXiv:2307.10485},
  year={2023}
}

@article{iaroshev2024evaluating,
  title={Evaluating Retrieval-Augmented Generation Models for Financial Report Question and Answering.},
  author={Iaroshev, Ivan and Pillai, Ramalingam and Vaglietti, Leandro and Hanne, Thomas},
  journal={Applied Sciences (2076-3417)},
  volume={14},
  number={20},
  year={2024},
  doi= {10.3390/app14209318},
  url= {https://doi.org/10.3390/app14209318}
}

@article{shah2024multi,
  title={Multi-Document Financial Question Answering using LLMs},
  author={Shah, Shalin and Ryali, Srikanth and Venkatesh, Ramasubbu},
  journal={arXiv preprint arXiv:2411.07264},
  year={2024}
}

@article{chen2024moa,
  title={Moa is all you need: Building llm research team using mixture of agents},
  author={Chen, Sandy and Zeng, Leqi and Raghunathan, Abhinav and Huang, Flora and Kim, Terrence C},
  journal={arXiv preprint arXiv:2409.07487},
  year={2024}
}

@article{yang2024finrobot,
  title={FinRobot: an open-source AI agent platform for financial applications using large language models},
  author={Yang, Hongyang and Zhang, Boyu and Wang, Neng and Guo, Cheng and Zhang, Xiaoli and Lin, Likun and Wang, Junlin and Zhou, Tianyu and Guan, Mao and Zhang, Runjia and others},
  journal={arXiv preprint arXiv:2405.14767},
  year={2024}
}

@article{zhou2024finrobot,
  title={FinRobot: AI Agent for Equity Research and Valuation with Large Language Models},
  author={Zhou, Tianyu and Wang, Pinqiao and Wu, Yilin and Yang, Hongyang},
  journal={arXiv preprint arXiv:2411.08804},
  year={2024}
}

@article{lewis2020retrieval,
  title={Retrieval-augmented generation for knowledge-intensive nlp tasks},
  author={Lewis, Patrick and Perez, Ethan and Piktus, Aleksandra and Petroni, Fabio and Karpukhin, Vladimir and Goyal, Naman and K{\"u}ttler, Heinrich and Lewis, Mike and Yih, Wen-tau and Rockt{\"a}schel, Tim and others},
  journal={Advances in neural information processing systems},
  volume={33},
  pages={9459--9474},
  year={2020}
}

@article{gao2023retrieval,
  title={Retrieval-augmented generation for large language models: A survey},
  author={Gao, Yunfan and Xiong, Yun and Gao, Xinyu and Jia, Kangxiang and Pan, Jinliu and Bi, Yuxi and Dai, Yi and Sun, Jiawei and Wang, Haofen and Wang, Haofen},
  journal={arXiv preprint arXiv:2312.10997},
  volume={2},
  year={2023}
}

@article{araci2019finbert,
  title={Finbert: Financial sentiment analysis with pre-trained language models},
  author={Araci, Dogu},
  journal={arXiv preprint arXiv:1908.10063},
  year={2019}
}

@article{yepes2024financial,
  title={Financial report chunking for effective retrieval augmented generation},
  author={Yepes, Antonio Jimeno and You, Yao and Milczek, Jan and Laverde, Sebastian and Li, Renyu},
  journal={arXiv preprint arXiv:2402.05131},
  year={2024}
}

@article{shneiderman2020human,
  title={Human-centered artificial intelligence: Reliable, safe \& trustworthy},
  author={Shneiderman, Ben},
  journal={International Journal of Human--Computer Interaction},
  volume={36},
  number={6},
  pages={495--504},
  year={2020},
  publisher={Taylor \& Francis}
}

@article{lee2004trust,
  title={Trust in automation: Designing for appropriate reliance},
  author={Lee, John D and See, Katrina A},
  journal={Human factors},
  volume={46},
  number={1},
  pages={50--80},
  year={2004},
  publisher={SAGE Publications Sage UK: London, England}
}

@article{islam2023financebench,
  title={Financebench: A new benchmark for financial question answering},
  author={Islam, Pranab and Kannappan, Anand and Kiela, Douwe and Qian, Rebecca and Scherrer, Nino and Vidgen, Bertie},
  journal={arXiv preprint arXiv:2311.11944},
  year={2023}
}

@inproceedings{sarmah2024hybridrag,
  title={Hybridrag: Integrating knowledge graphs and vector retrieval augmented generation for efficient information extraction},
  author={Sarmah, Bhaskarjit and Mehta, Dhagash and Hall, Benika and Rao, Rohan and Patel, Sunil and Pasquali, Stefano},
  booktitle={Proceedings of the 5th ACM International Conference on AI in Finance},
  pages={608--616},
  year={2024}
}

@article{lee2024multireranker,
  title={Multi-Reranker: Maximizing performance of retrieval-augmented generation in the FinanceRAG challenge},
  author={Lee, Joohyun and Roh, Minji},
  journal={arXiv preprint arXiv:2411.16732},
  year={2024}
}

@inproceedings{poliakov2024metadata,
  title={Multi-meta-rag: Improving rag for multi-hop queries using database filtering with llm-extracted metadata},
  author={Poliakov, Mykhailo and Shvai, Nadiya},
  booktitle={International Conference on Information and Communication Technologies in Education, Research, and Industrial Applications},
  pages={334--342},
  year={2024},
  organization={Springer}
}

@book{saunders2023researchframework,
  title={Research Methods for Business Students},
  author={Saunders, Mark and Lewis, Philip and Thornhill, Adrian},
  year={2023},
  edition={9},
  publisher={Pearson Education}
}

@article{sein2011actiondesign,
  title={Action design research},
  author={Sein, Maung K and Henfridsson, Ola and Purao, Sandeep and Rossi, Matti and Lindgren, Rikard},
  journal={MIS quarterly},
  pages={37--56},
  year={2011},
  publisher={JSTOR}
}

@article{tan2024structuredInput,
  title={Struct-X: Enhancing Large Language Models Reasoning with Structured Data},
  author={Tan, Xiaoyu and Wang, Haoyu and Qiu, Xihe and Cheng, Yuan and Xu, Yinghui and Chu, Wei and Qi, Yuan},
  journal={arXiv preprint arXiv:2407.12522},
  year={2024}
}

@article{auer2024docling,
  title={Docling Technical Report},
  author={Auer, Christoph and Lysak, Maksym and Nassar, Ahmed and Dolfi, Michele and Livathinos, Nikolaos and Vagenas, Panos and Ramis, Cesar Berrospi and Omenetti, Matteo and Lindlbauer, Fabian and Dinkla, Kasper and others},
  journal={arXiv preprint arXiv:2408.09869},
  year={2024}
}

@inproceedings{pfitzmann2022doclaynet,
  title={Doclaynet: A large human-annotated dataset for document-layout segmentation},
  author={Pfitzmann, Birgit and Auer, Christoph and Dolfi, Michele and Nassar, Ahmed S and Staar, Peter},
  booktitle={Proceedings of the 28th ACM SIGKDD conference on knowledge discovery and data mining},
  pages={3743--3751},
  year={2022}
}

@inproceedings{nassar2022tableformer,
  title={Tableformer: Table structure understanding with transformers},
  author={Nassar, Ahmed and Livathinos, Nikolaos and Lysak, Maksym and Staar, Peter},
  booktitle={Proceedings of the IEEE/CVF Conference on Computer Vision and Pattern Recognition},
  pages={4614--4623},
  year={2022}
}

@article{Leng2024LongCRA,
  title={Long Context RAG Performance of Large Language Models},
  author={Quinn Leng and Jacob Portes and Sam Havens and Matei A. Zaharia and Michael Carbin},
  journal={ArXiv},
  year={2024},
  volume={abs/2411.03538},
  url={https://api.semanticscholar.org/CorpusId:273849918}
}

@misc{rafiq2024ragie,
  title        = {How Ragie Outperformed the FinanceBench Test},
  author       = {Rafiq, Mohammed},
  howpublished = {\url{https://www.ragie.ai/blog/ragie-outperformed-financebench}},
  journal      = {Ragie Engineering Blog},
  publisher    = {Ragie},
  year         = {2024},
  month        = {Oct},
  day          = {22},
  note         = {Blog post}
}

@misc{rafiq2024ragie2,
  title        = {How Ragie Outperformed the FinanceBench Test — Part 2},
  author       = {Rafiq, Mohammed},
  howpublished = {\url{https://www.ragie.ai/blog/how-ragie-outperformed-the-financebench-test-part-2}},
  journal      = {Ragie Engineering Blog},
  publisher    = {Ragie},
  year         = {2024},
  month        = {Nov},
  day          = {25},
  note         = {Blog post, Part 2}
}

@misc{vectify2025mafingithub,
  title        = {FinanceBench Performance of Mafin 2.5},
  author       = {Vectify AI},
  howpublished = {\url{https://github.com/VectifyAI/Mafin2.5-FinanceBench}},
  year         = {2025},
  note         = {GitHub repository}
}

@misc{ray_srinivasan_2024amazon,
  author = {Ray, Shayan and Srinivasan, Bharathi},
  title  = {Reducing hallucinations in large language models with custom intervention using Amazon Bedrock Agents},
  year   = {2024},
  url    = {https://aws.amazon.com/blogs/machine-learning/reducing-hallucinations-in-large-language-models-with-custom-intervention-using-amazon-bedrock-agents/},
  note   = {Amazon Web Services Blog, published November 26, 2024}
}

@article{merrick2024arctic,
  title={Arctic-embed: Scalable, efficient, and accurate text embedding models},
  author={Merrick, Luke and Xu, Danmei and Nuti, Gaurav and Campos, Daniel},
  journal={arXiv preprint arXiv:2405.05374},
  year={2024}
}

@article{yu2024arctic,
  title={Arctic-Embed 2.0: Multilingual Retrieval Without Compromise},
  author={Yu, Puxuan and Merrick, Luke and Nuti, Gaurav and Campos, Daniel},
  journal={arXiv preprint arXiv:2412.04506},
  year={2024}
}

@article{enevoldsen2025mmtebmassivemultilingualtext,
    title={MMTEB: Massive Multilingual Text Embedding Benchmark},
    author={Kenneth Enevoldsen and others},
    publisher = {arXiv},
    journal={arXiv preprint arXiv:2502.13595},
    year={2025},
    url={https://arxiv.org/abs/2502.13595},
    doi = {10.48550/arXiv.2502.13595},
}

@inproceedings{wang-etal-2023-plan,
    title = "Plan-and-Solve Prompting: Improving Zero-Shot Chain-of-Thought Reasoning by Large Language Models",
    author = "Wang, Lei  and
      Xu, Wanyu  and
      Lan, Yihuai  and
      Hu, Zhiqiang  and
      Lan, Yunshi  and
      Lee, Roy Ka-Wei  and
      Lim, Ee-Peng",
    editor = "Rogers, Anna  and
      Boyd-Graber, Jordan  and
      Okazaki, Naoaki",
    booktitle = "Proceedings of the 61st Annual Meeting of the Association for Computational Linguistics (Volume 1: Long Papers)",
    month = jul,
    year = "2023",
    address = "Toronto, Canada",
    publisher = "Association for Computational Linguistics",
    url = "https://aclanthology.org/2023.acl-long.147/",
    doi = "10.18653/v1/2023.acl-long.147",
    pages = "2609--2634"
}

@inproceedings{yao2023react,
  title={React: Synergizing reasoning and acting in language models},
  author={Yao, Shunyu and Zhao, Jeffrey and Yu, Dian and Du, Nan and Shafran, Izhak and Narasimhan, Karthik and Cao, Yuan},
  booktitle={International Conference on Learning Representations (ICLR)},
  year={2023}
}

@article{erdogan2025plan,
  title={Plan-and-act: Improving planning of agents for long-horizon tasks},
  author={Erdogan, Lutfi Eren and Lee, Nicholas and Kim, Sehoon and Moon, Suhong and Furuta, Hiroki and Anumanchipalli, Gopala and Keutzer, Kurt and Gholami, Amir},
  journal={arXiv preprint arXiv:2503.09572},
  year={2025}
}

@misc{colvin_2025_15174950,
  author       = {Colvin, Samuel and
                  others},
  title        = {Pydantic},
  month        = apr,
  year         = 2025,
  publisher    = {Zenodo},
  version      = {v2.11.3},
  doi          = {10.5281/zenodo.15174950}}

@article{kim2025optimizingRag,
  title={Optimizing Retrieval Strategies for Financial Question Answering Documents in Retrieval-Augmented Generation Systems},
  author={Kim, Sejong and Song, Hyunseo and Seo, Hyunwoo and Kim, Hyunjun},
  journal={arXiv preprint arXiv:2503.15191},
  year={2025}
}

@article{gu2024survey,
  title={A survey on llm-as-a-judge},
  author={Gu, Jiawei and Jiang, Xuhui and Shi, Zhichao and Tan, Hexiang and Zhai, Xuehao and Xu, Chengjin and Li, Wei and Shen, Yinghan and Ma, Shengjie and Liu, Honghao and others},
  journal={arXiv preprint arXiv:2411.15594},
  year={2024}
}

@misc{Anthropic_2025, 
  title={The “think” tool: Enabling claude to stop and think}, 
  url={https://www.anthropic.com/engineering/claude-think-tool}, 
  publisher={Engineering at Anthropic}, 
  author={Anthropic}, 
  year={2025}, 
  month={Mar}
}

@article{choi2025finder,
  title={FinDER: Financial Dataset for Question Answering and Evaluating Retrieval-Augmented Generation},
  author={Choi, Chanyeol and Kwon, Jihoon and Ha, Jaeseon and Choi, Hojun and Kim, Chaewoon and Lee, Yongjae and Sohn, Jy-yong and Lopez-Lira, Alejandro},
  journal={arXiv preprint arXiv:2504.15800},
  year={2025}
}

@article{Zheng2023JudgingLWA,
  title={Judging LLM-as-a-judge with MT-Bench and Chatbot Arena},
  author={Lianmin Zheng and Wei-Lin Chiang and Ying Sheng and Siyuan Zhuang and Zhanghao Wu and Yonghao Zhuang and Zi Lin and Zhuohan Li and Dacheng Li and E. Xing and Haotong Zhang and Joseph E. Gonzalez and Ion Stoica},
  journal={ArXiv},
  year={2023},
  volume={abs/2306.05685},
  url={https://arxiv.org/pdf/2306.05685.pdf}
}

@inproceedings{Xu2024PrideAPA,
  title={Pride and Prejudice: LLM Amplifies Self-Bias in Self-Refinement},
  author={Wenda Xu and Guanglei Zhu and Xuandong Zhao and Liangming Pan and Lei Li and W. Wang},
  booktitle={Annual Meeting of the Association for Computational Linguistics},
  year={2024},
  url={https://arxiv.org/pdf/2402.11436.pdf}
}

@article{ye2024justiceprejudicequantifyingbiases,
      title={Justice or Prejudice? Quantifying Biases in LLM-as-a-Judge},
      author={Jiayi Ye and Yanbo Wang and Yue Huang and Dongping Chen and Qihui Zhang and Nuno Moniz and Tian Gao and Werner Geyer and Chao Huang and Pin-Yu Chen and Nitesh V Chawla and Xiangliang Zhang},
      journal={arXiv preprint arXiv:2410.02736},
      year={2024}
}

@article{khattab2023dspy,
  title={Dspy: Compiling declarative language model calls into self-improving pipelines},
  author={Khattab, Omar and Singhvi, Arnav and Maheshwari, Paridhi and Zhang, Zhiyuan and Santhanam, Keshav and Vardhamanan, Sri and Haq, Saiful and Sharma, Ashutosh and Joshi, Thomas T and Moazam, Hanna and others},
  journal={arXiv preprint arXiv:2310.03714},
  year={2023}
}

@inproceedings{sarmah2023towards,
  title={Towards reducing hallucination in extracting information from financial reports using large language models},
  author={Sarmah, Bhaskarjit and Mehta, Dhagash and Pasquali, Stefano and Zhu, Tianjie},
  booktitle={Proceedings of the Third International Conference on AI-ML Systems},
  pages={1--5},
  year={2023}
}

@article{lai2024sec,
  title={SEC-QA: A Systematic Evaluation Corpus for Financial QA},
  author={Lai, Viet Dac and Krumdick, Michael and Lovering, Charles and Reddy, Varshini and Schmidt, Craig and Tanner, Chris},
  journal={arXiv preprint arXiv:2406.14394},
  year={2024}
}

@article{Taghvaei2024CombiningFDA,
  title={Combining Financial Data and News Articles for Stock Price Movement Prediction Using Large Language Models},
  author={Fatemeh Taghvaei and Ali Elahi},
  journal={2024 IEEE International Conference on Big Data (BigData)},
  year={2024},
  pages={4875-4883},
  url={https://api.semanticscholar.org/CorpusId:273812246}
}

@misc{europarl2023AIAct,
  author       = "{European Parliament}",
  title        = "{EU AI Act: first regulation on artificial intelligence}",
  year         = "2025",
  url          = "https://www.europarl.europa.eu/topics/en/article/20230601STO93804/eu-ai-act-first-regulation-on-artificial-intelligence",
  urldate      = "2025-05-12"
}

@article{nguyen2024enhancing,
  title={Enhancing Q\&A with Domain-Specific Fine-Tuning and Iterative Reasoning: A Comparative Study},
  author={Nguyen, Zooey and Annunziata, Anthony and Luong, Vinh and Dinh, Sang and Le, Quynh and Ha, Anh Hai and Le, Chanh and Phan, Hong An and Raghavan, Shruti and Nguyen, Christopher},
  journal={arXiv preprint arXiv:2404.11792},
  year={2024}
}

@article{tang2025finmteb,
  title={FinMTEB: Finance Massive Text Embedding Benchmark},
  author={Tang, Yixuan and Yang, Yi},
  journal={arXiv preprint arXiv:2502.10990},
  year={2025}
}

@misc{Anthropic_2024, 
  title={Building effective AI agents}, 
  url={https://www.anthropic.com/engineering/building-effective-agents}, 
  journal={Building Effective AI Agents \ Anthropic}, 
  publisher={Engineering at Anthropic}, 
  author={Anthropic}, 
  year={2024}, 
  month={Dec}
}

@misc{Zhang_HowWeBuildEffectiveAgents_2025,
    author       = {Zhang, Barry},
    title        = {How We Build Effective Agents},
    year         = {2025},
    publisher    = {AI Engineer},
    howpublished = {YouTube},
    note         = {Presented at the AI Engineer Summit 2025, New York. [Accessed: May 14, 2025]},
    url          = {https://www.youtube.com/watch?v=051042_2p5E}
}
\end{document}